\documentclass[a4paper]{mycls_preview} % for no journal information

\usepackage[numbers]{natbib}
\usepackage{url}
\usepackage{svg}
\usepackage{pdfpages}
\jname{Xxxx. Xxx. Xxx. Xxx.}
\jvol{AA}
\jyear{YYYY}
\doi{10.1146/((please add article doi))}

\begin{document}

% Page header
\markboth{B\"utepage et al.}{Human-Robot Collaboration: From Psychology to Social Robotics}

% Title
\title{ Human-Robot Collaboration: From Psychology to Social Robotics}

%Authors, affiliations address.
\author{Judith B\"utepage,  Danica Kragic
\affil{Robotics, Perception and Learning Lab (RPL), CSC, KTH Royal Institute of Technology, Stockholm, Sweden, 114 28; email: butepage$|$dani@kth.se}}

%Abstract
\begin{abstract}
With the advances in robotic technology, research in human-robot collaboration (HRC) has gained in importance. For robots to interact with humans autonomously they need active decision making that takes human partners into account. However, state-of-the-art research in HRC does often assume a leader-follower division, in which one agent leads the interaction. We believe that this is caused by the lack of a reliable representation of the human and the environment to allow autonomous decision making. This problem can be overcome by an embodied approach to HRC which is inspired by psychological studies of human-human interaction (HHI). In this survey, we review neuroscientific and psychological findings of the sensorimotor patterns that govern HHI and view them in a robotics context. Additionally, we study the advances  made by the robotic community into the direction of embodied HRC. We focus on the mechanisms that are required for active, physical human-robot collaboration. Finally, we discuss the similarities and differences in the two fields of study which pinpoint directions of future research. 
\end{abstract}

%Keywords, etc.
\maketitle

%Table of Contents
\tableofcontents

\begingroup
 
%% main text
\section{INTRODUCTION}
\label{sec:intro}

% how embodiment came into robotics
% how smcs seem to govern hhi
% how to take that knowledge and transfer it to hri

% one paragrapgh about pHRC and why a SMC approach is awesome
Within the last decades, human-robot interaction (HRI) has become an important field of robotics. As robots and humans share a growing number of workspaces, the development of safe and intuitive systems is of high importance. The physical proximity of interacting partners poses the challenge to integrate human actions into the robot's decision making process. Commonly, this challenge is avoided by following the master-slave principle, i.e. to design a compliant robotic system that reacts towards human actions and merely assists the human in achieving a goal \cite{jarrasse14}. Many interaction scenarios in everyday life however call for actively collaborating partners that can autonomously contribute to a shared task. 

\begin{figure}[h!]
\centering
\includegraphics [ width= 0.9 \textwidth]{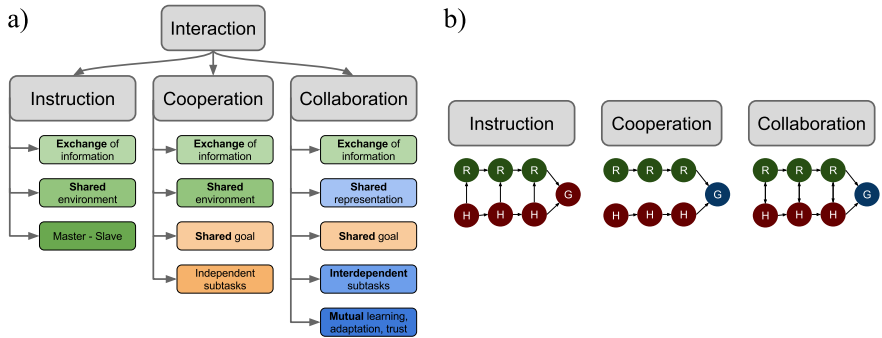}
\caption{a) Three types of interaction between humans and robots, increasing in the amount of interdependency. b) The interaction between the robot (R, green) and human (H, red) to reach the human's (G,red) or shared (G, blue) goal over time. Arrows indicate dependencies. } \label{Fig:introinteraction}
\end{figure}

We can view the differences in the level of interdependence and shared representations as enlisted in \textbf{Figure \ref{Fig:introinteraction}a)}.  Interaction between humans and robots can be characterized as either instruction, cooperation or collaboration. \textbf{Figure \ref{Fig:introinteraction} b)} depicts the dependencies of the robot and human in a goal-directed interaction over time for each of these three categories.

We define instruction as a sequence of actions in a HRI setting which is governed by the human's decision making, i.e. we view the human as the coordinator who influences the robot as shown in the left part of \textbf{Figure \ref{Fig:introinteraction}b)}. While the interaction relies on an exchange of information between the partners in a shared environment, the leader-follower setting prevents advanced social interaction. A common example of this interaction is learning from demonstration \cite{argall09}.

Cooperation on the other hand is defined as a sequence of actions in a HRI setting towards a shared goal. However, each partner is working independently towards separate subtasks, as shown in the middle part of \textbf{Figure \ref{Fig:introinteraction}b)} . Once the subtasks are divided between partners, their actions are independent of each other.

Finally, collaboration requires shared representations of the environment. We define collaboration as a sequence of interdependent actions in a HRI setting towards a shared goal. The right part of \textbf{Figure \ref{Fig:introinteraction}c)} represents this interdependency by mutual influences over time. Collaboration is the basis for mutual learning and mutual adaptation and requires mutual trust. 

In HRI, cooperation and collaboration are often used interchangeably, even though they are conceptually different \cite{keast2007getting}. The differences between cooperation and collaboration can be high-lighted with an example. In a joint cooking scenario, the task of preparing a meal could be divided into subtasks, preparing the sauce, the vegetables, etc., which can be executed independently while working towards a shared goal. Collaboration occurs when the partners adapt to each other, when they track each other's intentions and needs and understand when to pitch in. For example, cutting onions not only for one's own task but also for the partner shows that one actively engages in the activities of the partner. 

An equal role distribution as required by collaboration results in more efficient interactions and a reduced cognitive workload for the human partner. 
This is especially prevalent in physical interactions, i.e. scenarios with direct physical contact or physical contact mediated through an object of shared interest. Physical interactions often require interdependent coordination of actions based on a shared representation of the task and the environment. However, the majority of state-of-the-art studies in physical human-robot interaction assumes an instruction or cooperation approach.
Thus, we see the need for a shift from HRI to  human-robot collaboration HRC. 
 
In order to design collaborative robots that act intuitively for the human partner, the following aspects need to be taken into consideration:

\begin{enumerate}
\item Which principles and mechanisms govern HHI that can inspire the design of  HRC systems?
\item What kind of state representations and mechanisms are optimal for HRC? 
\item How can we achieve autonomous, collaborative behavior?
\end{enumerate}

\begin{marginnote}[]
\entry{Sensorimotor contigencies}{ Sensorimotor contigencies describe the correlations between actions and the perceptual change they induce. In this view, cognition does not require abstract world models to enable actions. Instead the need for cognition is driven by the need for action and cognitive processes are embedded in sensorimotor processes \cite{engel2013s}.}
\end{marginnote}
In this survey, we aim at developing a conceptual approach towards HRC which connects the above posted questions. We hypothesize that HRC can benefit from ideas of embodied intelligence and sensorimotor contingencies (SMCs).

Firstly, a sensorimotor approach towards non-verbal social interaction is supported by findings in psychology and neuroscience \cite{dipaolo12} (in accordance with question 1).  Secondly, sensorimotor representations are highly beneficial for any kind of HRC as the human is automatically represented in the robot's sensorimotor spaces without the need for explicit modeling (in accordance with question 2). Finally, a shared representation in terms of sensorimotor signals can facilitate autonomous action selection in collaborative settings (in accordance with question 3).  Taking this viewpoint on social cognition allows to move away from the need of a theory of mind and serves the development of naturally interacting robots \cite{butepage2016social}.

% aim of the survey
Due to the similarity of interests in both HHI and HRC, it is of importance to establish a coherent picture of social interaction. While HHI can benefit from this unification by using embodied agents as a testing ground for theoretical explanations, HRC needs to incorporate HHI research in order to create active, adaptive systems that can fluently interact with humans. The goal of this survey is  to review both experimental findings and theoretical concepts of HHI and HRC with a special emphasis on non-verbal interaction based on social sensorimotor contingencies (socSMCs). We aim at a systematic approach towards active HRC that does not rely on master-slave and teacher-learner principles. 

% short overview of different topics 
We begin this survey with a short discussion of a structured framework of socSMCs in the continuation of this section. An in-depth discussion of experimental results from HHI research follows in Section \ref{sec:humhum}. Subsequently,  in Section \ref{sec:humrob} we present  developments in HRC with a special focus on social sensorimotor signals and active collaboration. In order to give a coherent account of HHI and HRC, we will point out both important similarities and gaps between the two approaches in Section \ref{sec:future}. Finally, we will discuss future research directions within the topic of socSMCs and social interaction that can pave the way to a deeper understanding of  social cognition.

\subsection{\textbf{From embodied intelligence to social cognition}}
\label{sec:overview}
% embodied intelligence
Since the second half of the 1980, the concept of embodied intelligence has revolutionized  artificial intelligence \cite{brooks91}. Instead of logical architectures and knowledge representation, the embodied view argues that intelligent behavior emerges naturally from the interplay between motor and sensory channels \cite{oregan01}, as depicted in \textbf{Figure \ref{Fig:smc}}. This coupling between an agent and its environment through sensorimotor signals and constant inference and feedback loops is suggested to account for complex behavior without the need of high-level reasoning.  In this view, internal representations become obsolete because the environment is its own representation that is actively sampled via sensory channels and manipulated by self-induced actions. Physical laws constrain the action possibilities of any agent but can also be exploited to gain advantages, see \cite{pfeifer06} for an in-depth discussion.

\subsubsection{Sensorimotor contingencies}
Sensorimotor contingencies (SMCs) are statistical properties of action-perception loops that allow categorization of events, actions and objects and fluent interaction between an agent and its environment. Due to the high level of uncertainty and noise inherent to most biological systems,  sensorimotor learning and control has been cast into probabilistic frameworks \cite{wolpert07, orban11, haruno01}. This mathematical formulation makes SMCs highly accessible to robotics by providing tools for adaptive, cognitive systems.

SMCs are believed to be of high importance for perception and reactive behavior but also to contribute to high-level cognition. From object recognition  \cite{hogman13}  and manipulation \cite{Stork15} to path planning \cite{boots11}, many problem settings in robotics can be solved with help of action-perception-effect loops. Even more so, advances in neuroscience and psychology suggest that even social cognition is grounded in SMC signals \cite{wolpert03}.

\begin{figure}[t!]
\centering
\includegraphics [ width= 0.40 \textwidth]{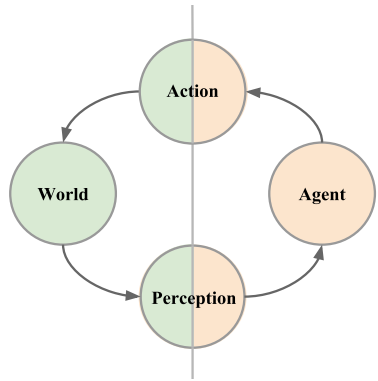}
\caption{Embodied intelligence denotes the idea that complex behavior arises due to a constant coupling between an agent and its environment through sensorimotor channels. } \label{Fig:smc}
\end{figure}

% biologically inspired
Concepts in embodied intelligence are highly influenced by biological systems that naturally represent and incorporate these mechanisms. While early focus lay on emerging patterns such as reflexes and simple heuristics as found in insects \cite{maris96}, later studies started to aim at human-like systems \cite{montesano08}. In order to create such a system, the understanding of sensorimotor processes in humans is substantial. %Thus, the embodied approach to artificial cognition relies on psychological and neuroscientific findings that support and inspire robotic system design. 

\subsubsection{Towards a multi-agent approach}
% study more than one agent - human
In a historical perspective,  research in psychology and neuroscience on one hand and research in artificial intelligence and robotics on the other hand has mainly concentrated on the individual agent in isolation. 
This study of a single embodied subject carries its own advantages. Nevertheless, we cannot neglect that a substantial amount of behavioral and sensorimotor patterns, emerging in HHI settings, cannot be reduced to  individual accounts \cite{loehr201313}. Instead, the dyadic configuration gives rise to complex structures requiring a shift of  focus in research. Similarly, the implementation of an artificial, embodied agent, that is able to actively cooperate with humans, needs to incorporate knowledge about these interaction dynamics governing HHI.

% robot : sensorimotor
The study of artificial, interactive systems can be divided into two main streams of research. On the one hand, we have language-based agents such as dialog systems rooted in the natural language processing community. In this context, language as a means of communication and sign of intelligence has received a lot of attention, see e.g. \cite{jokinen09, uthus13}.  Indeed, the Turing Test proposes that a system capable of indistinguishably imitating  a human conversation partner can be taken as a first step towards artificial   intelligence \cite{turing50}. However, a focus on language disregards that a high amount of communication between humans is non-verbal \cite{mehrabian71}. Instead, gestures, facial expressions and gaze embedded in mutual prediction and signaling mechanisms convey instant information exchange. Recently,  HRC  has gained interest in the implementation of these embodied social signaling techniques, see e.g. \cite{dragan14, vitale14, Moon14,  mainprice15}. HRC benefits from this sensorimotor approach towards social interaction since it allows to encode interaction variables in terms of body-internal signals which reduces the requirement for complex representations. On these grounds, the robot can act autonomously in HRC settings by moving from the master-slave principle towards equal role sharing.

\subsection{\textbf{Social sensorimotor contingencies}}

In order to present socSMCs in a framework that allows interdisciplinary discussion, we follow ideas developed in the EU FET Proactive project "socSMCs" \cite{socSMCproject} and divide known phenomena in interaction into three categories: check SMCs, sync SMCs and unite SMCs. These concepts are shortly introduced below and will guide the following review of literature in HHI and HRC.  

\begin{figure}[b!]
\centering
\includegraphics[width=1\textwidth]{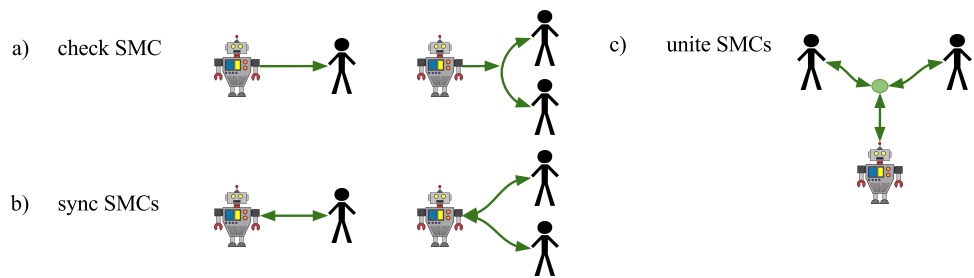}
\caption{The three categories of social sensorimotor contingencies - check SMCs, sync SMCs and unit SMCs - representing different levels of interaction.} \label{Fig:socsmcs}
\end{figure}

\subsubsection{check SMCs}
% Check SMCs involve unidirectional coupling, one
% agent predicting another agent’s actions or the interaction between two other
% agents. Behaviourally, this may lead e.g. to entrainment of one agent to a group
% of other agents. 
Check SMCs denote unidirectional prediction by one agent of another agent's actions or the interaction between two other agents as shown in \textbf{Figure \ref{Fig:socsmcs}a)}  on the left and right respectively. This implies a coupling between the observing agent's perceptual channels and the motor behavior of other agents. Effectively, check SMCs may lead to entrainment of the observing agent i.e. synchronization or reciprocal behavior such as imitation. 

\begin{marginnote}[]
\entry{Sensorimotor coupling}{Sensorimotor coupling describes the synergy of sensory and motor processes. Through adaptation of these processes, sensory signals are associated with motor outputs and vice versa. }
\end{marginnote}

\subsubsection{sync SMCs}
% Synch SMCs enable bidirectional coupling, both agents mutually
% predicting each other. This may lead to mutual entrainment, allowing cooperation,
% joint attention, turn-taking, and shared action goals. 
In contrast to check SMCs, sync SMCs imply mutual prediction and coupling of two agents as shown in \textbf{Figure \ref{Fig:socsmcs}b)}. This allows for high-level coordination such as joint attention, turn-taking and collaboration towards shared action goals. For sync SMCs the agents are required to represent each other in terms of sensorimotor loops and to consider each other's actions when making decisions.

\subsubsection{unite SMCs}
% Unite SMCs promote
% group-related, multidirectional coupling. They express a coupling mode characterized
% by emergence of higher-order interaction patterns that cannot fully be
% explained by the bivariate interactions among the group members. This may allow
% the emergence of group mental states, group habits, and group emotions.
Finally, unite SMCs describe group phenomena that cannot be explained by dyadic interactions alone as depicted in \textbf{Figure \ref{Fig:socsmcs}c)} . Group habits, group mental states and emotions fall under this category. While check SMCs and sync SMCs are of high importance for HRC, unite SMCs are an additional component that carries less importance in the direct interaction.

\section{HUMAN-HUMAN INTERACTION}
\label{sec:humhum}
% -----------------------------------------------------------------------------
% Intro to HHI
% -----------------------------------------------------------------------------
Up to this point we have discussed the general ideas behind embodied intelligence in a social context. In order to understand how these principles can be implemented in HRC, the underlying mechanisms in human interaction need to be understood. In this section we will review neuroscientific and psychological findings that fall under the umbrella of socSMCs and point out their importance for robotic research.  

The subject of intelligence has fascinated humankind for millennia. The ability to reason logically, memorize and gain new knowledge are highly respected in western culture. Since we associate intelligence with logic and abstract thinking, we only notice the importance of social intelligence when it is absent. For example, individuals who suffer from  neurodevelopmental disorders such as the Autistic Spectrum Disorder are highly impaired in their everyday life due to the lacking ability to read and send social signals \cite{caronna08}. The severity of this deficit highlights the significance of social cognition and the need to gather a deep understanding of its underlying dynamics. To the present day, it is unclear how we are able to learn and apply social laws effortlessly, how we are able to make complex inferences and predictions about others' future actions and how we can handle a  multitude of hidden variables. However, psychological and neuroscientific research has uncovered a number of important phenomena that influence joint action. %These phenomena can usually be categorized as either automatic coupling or planned coordination of human partners based on SMCs. 
We will begin this section by introducing fundamental aspects of sensorimotor communication in HHI in Section \ref{sec:socsmcjointaction}, followed by an in-depth discussion of check and sync SMCs phenomena in Section \ref{sec:checkhuman} and Sections \ref{sec:syncentrainhuman} and \ref{sec:syncplanhuman} respectively.  

% -----------------------------------------------------------------------------
% smcs in joint action
% -----------------------------------------------------------------------------
\subsection{\textbf{socSMCs - Sensorimotor signals in joint action}}
\label{sec:socsmcjointaction}

The difference between SMCs in an isolated environment and SMCs in a social context is the fact that one's actions do not only impact the environment and one's own perceptual state but can also influence the actions of the social partner. Additionally, others' actions change the environment and might lead to perceptual changes in the observing agent. These socSMCs are crucial in joint action where "two or more individuals coordinate their actions in space and time to bring about a change in the environment" \cite{sebanz2006joint}. A common example of joint action is two people carrying a heavy object together. In order to successfully move the object, the two need to coordinate their actions with help of a number of mechanisms which  will be discussed in detail in the following sections.

% multimodal
In interaction, multimodal sensory signals can be integrated to approximate the current state with higher precision than if presented in a single sensory channel \cite{wahn2015multisensory}. This suggests that attentional resources are well distributed across sensory channels. As an example, haptic and visual integration follows the principle of minimal variance, i.e. a maximum likelihood estimate \cite{ernst2002humans}, additional sensory channels can lower the uncertainty about the partner's actions. 

% monitoring -> what is the other doing right now
Additionally, humans  monitor their partner's actions and potential errors \cite{bates2005external} which is essential to react timely and to adapt to the state changes caused by others. 
% signalling and predicting
This is also facilitated by predicting others' future actions and the perceptual changes they might cause. Especially in joint action, prediction of actions and their spatial and temporal aspects is required for action planning and coordination, see reviews on this topic \cite{brown2012role} and \cite{sebanz2009prediction}. To elevate one's own readability and predictability, a common coordination strategy is to actively signal in terms of changes in kinematics, e.g. action profiles or velocities \cite{sacheli2013kinematics}. 
% emergent and planned

Finally, phenomena in joint action are often divided into low-level emergent coordination and high-level planned coordination. While in both areas, coordination arises effortlessly and mostly subconsciously, the main difference is that planned coordination facilitates a joint action towards a common goal \cite{Knoblich11} while emergent coordination is caused by a coupling of perceptual input and motor output processes \cite{richardson08}.

% socSMCs and robots
\subsubsection{A robotics viewpoint}
This short introduction indicates how research in HHI structures the problem into different mechanisms and hierarchical level. Following these structures could facilitate the classification and comparison of different research questions and methodologies developed within the robotics community. Therefore, we will review HRC studies related to socSMCs following the categories developed in the HHI research community in Section \ref{sec:humrob}.

% -----------------------------------------------------------------------------
% check Smcs - HHI
% -----------------------------------------------------------------------------
\subsection{\textbf{check SMCs - unidirectional coupling}}
\label{sec:checkhuman}

The unidirectionality of check SMCs entails that one agent couples its own sensorimotor system to the actions of another agent. In contrast to a bidirectional scenario such as collaboration or teacher-student settings the second agent does not actively adjust its actions with respect to the observing agent. Thus, unidirectional coupling or coordination is mostly concerned with the understanding and prediction of sensory changes induced by others' actions. In this section we discuss the biological foundation of these processes in Section \ref{sec:mirrorhum} and continue by presenting phenomena concerned with imitation and mimicry in Section \ref{sec:imitationhum}.

%----------------------------------------------------------------------------
\subsubsection{Understanding actions by simulation} 
\label{sec:mirrorhum}
% mirror neurons

A prominent theory of action understanding and prediction in the primate brain is action simulation which is also known as mirroring \cite{pezzulo13} or ideomotor theory. This approach takes the theory of common coding of action and perception \cite{prinz1990common} one step further to propose that the encoding of perceiving others' actions and the execution of our own actions share common brain structures. In this view, action understanding becomes a matter of action simulation in the brain and prediction of others' actions is performed by engaging one's own motor system \cite{van2013action}.  In terms of biological factors, this process is believed to be driven by mirror neurons \cite{rizzolatti04} which are active both during action execution and the perception of someone else executing the same action. In a broader sense, the mirror neuron system has been found to extend beyond object-related actions to encode intentions and social interactions \cite{Sliwa745}. While some studies argue that the mirror neuron system is genetically predefined \cite{oberman2007human}, a more recent direction suggests that it is the result of associative learning in a social context \cite{keysers14, cook2014mirror}. In terms of socSMCs, the simulation theory could account for unidirectional coupling as well as subconscious imitation and mimicry which will be discussed below.

% Understanding actions by simulation and robots
\paragraph{A robotics viewpoint}
Human action recognition from videos has been extensively studied in the computer vision community. Depending on the dataset the performance of these systems can vary between 40 \% and 95 \% \cite{wang2016robust}. However, these systems usually require a substantial amount of training data and they are confined to the actions contained in the data. For successful HRC, we require a system that can recognize a variety of actions, which might even be novel, in cluttered environments. Therefore, the idea of mapping human actions to the robot's own states in order to facilitate action understanding is appealing. This mapping is not trivial, due to the so called correspondence problem, which is discussed in Section \ref{sec:corrprob}.

%----------------------------------------------------------------------------
\subsubsection{Mimicry and imitation} 
\label{sec:imitationhum}
% Mimicry and imitation
Basic motor coordination between individuals occurs automatically and goes mostly unnoticed but can also be controlled behavior. Perceiving another person performing an action can result in involuntary mimicry, emulation or imitation \cite{lopes2010abstraction}. Mimicry is associated with pure action mirroring without considering the intention behind it. Emulation describes the reproduction of a goal-directed behavior without adapting the same set of actions leading to the goal. Finally, imitation combines action mirroring in a goal-directed context.  

A common example for mimicry is the yawning reflex which can be triggered by seeing or hearing someone else yawn  \cite{Platek2005}. This so-called Chameleon effect  \cite{chartrand99} implies automatic mirroring of facial expressions, gestures and body movement. It is under debate at which age human infants start mimicking and to which degree the brain is hard-wired for this behavior \cite{jones09}. 
Whether humans emulate or imitate a demonstrated action can depend on the context \cite{meltzoff1988infant} and the social background \cite{nielsen2006copying}. Imitation is not only influenced by whether a subject belongs to the same social group but also by whether the context is cooperative of competitive \cite{gleibs2016group}. The role of imitation for infant development, both for learning motor skills and to engage in social context, is profound \cite{over2013social}. For example, being imitated increases interpersonal trust \cite{over2013children}. Substantial deficits of imitation capabilities in  children diagnosed with autism spectrum disorder \cite{ingersoll08} indicate that involuntary imitation is highly coupled with social cognition. 

% Mimicry and imitation and robots
\paragraph{A robotics viewpoint}
Imitation of human actions or movements does usually serve as a learning mechanism in robotics. The human can be imitated on a high level, i.e. the effect of an action is emulated, or on a lower, more kinematic level \cite{lopes2010abstraction}. Learning in a social context has the advantage of skill acquisition without explicit definitions of cost functions. Different approaches towards the imitation problem are discussed in Section \ref{sec:affordance}.

% -----------------------------------------------------------------------------
% sync Smcs - HHI
% -----------------------------------------------------------------------------
\subsection{\textbf{sync SMCs - mutual coupling}}
\label{sec:syncentrainhuman}

Sync SMCs extend the notion of check SMCs by mutual coupling, i.e.  two agents align their actions in time and space. This does not only denote synchronous actions but any kind of reciprocal action-perception-effect loop that coordinates the interaction. In this section, we highlight the foundations of entrainment that arises in the context of   emergent coordination in Section \ref{sec:entrainmenthum}. Mutual coupling also affects coordination in interaction such as turn-taking, discussed in Section \ref{sec:turntakinghum} and the social perception of interaction partners, described in Section \ref{sec:effectshum}.    

%----------------------------------------------------------------------------
\subsubsection{Mutual Entrainment} 
\label{sec:entrainmenthum}
% entrainment
Entrainment commonly describes the coupling of rhythmic or oscillatory systems, i.e. that one rhythm couples to another in a systematic manner. In humans, a sensorimotor coupling between perceptual information about others' movements and one's own actions can cause subconscious adaptation and the emergence of synchrony \cite{richardson08}. As an example, phase-locking between the gait patterns of two persons walking next to each other is influenced by auditory and haptic information \cite{Zivotofsky12}. In experiments with rocking chairs, Richardson et al. \cite{richardson07} showed that partners rocking in near proximity to each other will synchronize voluntarily and involuntarily even if the natural rocking frequency of the chairs is different.

% Entrainment and robots
\paragraph{A robotics viewpoint}

Entrainment in robotics can be viewed from two directions. On the one hand, a robotic system can benefit from synchronization during motor learning due to reduced noise. On the other hand, therapeutic robotics requires robots that adapt their actions towards the patient so that the human partner can improve motor skills.
We can distinguish between pure motor entrainment, which results in reciprocal behavior, and adaptation of higher order cognitive states, which can result in a shared state representation.  Basic synchronization and entrainment in HRC is presented in Section \ref{sec:synchronizationrobot}. Mutual adaptation of robots and humans, which can go beyond these basic skills,  is discussed in Section \ref{sec:mutualadaprobot}.

%----------------------------------------------------------------------------
%turn taking
\subsubsection{Turn-taking} 
\label{sec:turntakinghum}

Turn-taking is a phenomenon in human interaction that is mostly associated with verbal dialogue \cite{sacks1974simplest}. Nevertheless, turn-taking can be observed in many sequential task settings such as joint pick-and-place tasks. As the turn-taking by the observing partner is often initiated before the  turn of the currently acting partner is over, it is believed that humans rely on a variety of non-verbal cues such as prosody and body motion \cite{duncan1972some}. Accurate timing and turn-taking behavior is governed by the ability to predict others' actions and to inhibit one's own actions \cite{meyer2015journal}. 

% Turn-taking and robots
\paragraph{A robotics viewpoint}
Turn-taking in a broader context includes not only dialogue, but also physical scenarios such as hand-overs and joint activities such as cooking or building a brick tower. In order to coordinate with a human, the robot needs to be able to signal when the human should take over and to understand when it is its own turn. As the human should not have to wait unnecessarily long for the robot to initiate an action, the robot needs to anticipate human actions and to interpret the non-verbal cues sent by its partner. We discuss general turn-taking in HRC in Section \ref{sec:turntakingrobot}.  Furthermore, we review work in predicting of and signaling to a human in Section \ref{secsec:prediction_robot}  and Section \ref{sec:signalrobots} respectively.

%----------------------------------------------------------------------------
\subsubsection{Effects of emergent coordination} 
\label{sec:effectshum}
% Effects of emergent coordination
When discussing emergent coordination in human interaction, the question arises which function these phenomena serve. Both cognitive and emotional benefits have been investigated. For instance, in-phase synchronization during a conversation facilitates the memory of facial expressions and utterances \cite{macrae08}. Thus, executive functions such as attention, memory and learning are enhanced by coordinated behavior \cite{chartrand99, keller14}. More important for the social interaction are the emotional effects. Belonging to the same social group is often followed by mutual mimicry and synchronization while imitation by an individual of another social group is perceived negatively. By this, group membership  and the feeling of connectedness is enhanced \cite{van09, marsh09}. A higher level of coordination also increases out willingness to cooperate \cite{wiltermuth09}.

% -----------------------------------------------------------------------------
% sync Smcs - HHI
% -----------------------------------------------------------------------------
\subsection{\textbf{sync SMCs - planned coordination}}
\label{sec:syncplanhuman}

% short intro to sync
While emergent coordination is usually described independent of the interaction goal, planned coordination arises in a joint task setting. This implies that interaction partners might work towards a shared goal and requires a representation of the partner's action space and high-level coordination of actions. Here we discuss those aspects of planned coordination which seem important with respect to HRC: shared representations (Section \ref{sec:sharedrephuman}), prediction (Section \ref{sec:predictionhum}) and signaling (Section \ref{sec:signalinghum}). For deeper discussions, we refer to reviews on this topic such as  \cite{sebanz2009prediction}  and  \cite{Knoblich11}.

%----------------------------------------------------------------------------
\subsubsection{Shared representations} 
\label{sec:sharedrephuman}

Shared sensorimotor representations build the foundation for cooperative and collaborative behavior in humans. They allow to communicate on a common ground and to automatically engage in joint action. In this section, we list several aspects that rely on a shared representation.  
In terms of sync SMCs, shared representations emphasize the automatic sensorimotor coupling between interaction partners during planned coordination. A joint setting imposes structure on a task which can change the sensorimotor representation and planning of both partners.

% Representing others
\noindent \textbf{Representing others}
The representation of others in joint action implies a neural correlate of their actions that influences the actions of the observer. 
For example, Sebanz et al. \cite{sebanz03} showed that others' actions influence one's own actions even if the observations are not relevant for task completion. Additionally, the  presence or prospect of a potential interaction partner changes the sensorimotor perception of a task. 
Participants judge a load to be lighter when it is to be carried with a partner \cite{doerrfeld2012expecting} and switch between acting on intrapersonal and interpersonal affordances based on a joint affordance space \cite{Richardson07a}. Moreover, not only the lowering of cost in a joint setting influences performance judgments but also the additional cognitive load of having to coordinate with a partner \cite{meagher2014costs}.

\begin{marginnote}[]
\entry{Affordances}{Affordances describe the action possibilities that objects offer to an agent.}
\end{marginnote}

% Representing tasks
\noindent \textbf{Shared task representation}
% tasks
The representation of others suggests that the perception-action-effect loop might not be a purely associative mechanism but be influenced by higher level structures such as social engagement or the task. For example, subjects seem to form different sensorimotor representations during imitative vs. complementary tasks \cite{van2008understanding, poljac2009understanding} suggesting that the social structure of a task influences its representation.

% Representing others
\noindent \textbf{Mutual adaptation}
In the absence of language and sensory connections when facing a cooperative task, a common symbolic language emerges spontaneously between subjects \cite{galantucci2005experimental}. Thus, in the absence of a shared representation, humans readily and automatically adapt their representation mutually in joint action to reach a shared goal. 

% Representing others
\noindent \textbf{Shared intentions}
Sharing representations includes also the sharing of intentions. According to Tomasello et al. \cite{tomasello2005understanding} intentions incorporate both the goal of an action and the action plan. Thus, shared intentions between individuals imply a common goal and a coordinated action plan that all partners have agreed upon. The ability to share intentions develops during infancy and is vital for any cooperative or collaborative interaction. They govern our ability to coordinate role division and to select actions during joint action.

% Representation and robots
\paragraph{A robotics viewpoint}

One of the challenges in human-robot collaboration is the question of how to represent and integrate human partners into the decision making of the robot. This differs from common robotic tasks as the human is an independent agent that can introduce nearly arbitrary changes into the environment. In order to create a shared representation of the human and robotic partner, the robot needs to be able to represent the human within the environment, as discussed in Section \ref{sec:representationrobot}. Additionally, it needs to be able to infer the human's intentions and to plan in collaborative settings, which we review in Sections  \ref{sec:intentioninferencerobot} and \ref{sec:collaborativetaskplanningrobot} respectively.

%----------------------------------------------------------------------------
\subsubsection{Prediction and anticipation}
\label{sec:predictionhum}
% predicting on the motor level and the high level action level in a non-interaction scenario

When engaging with the environment, humans are believed to make use of sensorimotor forward models. These models predict the sensory change that is induced by an action given the current state \cite{wolpert1995internal}. The error between prediction and actual outcome can drive motor learning and adaptation. Similarly, prediction of others' actions is vital for a fluent physical interaction and for making inferences over others' goals. Following the ideas discussed above, the mirror neuron system might drive these predictions and open the connection to intentions behind actions \cite{kilner07}. Prediction is involved in both check and sync SMCs. On the one hand, it does only involve the observing agent. On the other hand, the function of this process is to be able to take the actions of one's partner into account; indirectly enabling mutual syncing of sensorimotor processes. 

On a higher level, the prediction of the timing, the location and the nature of actions has been studied in the context of social interaction  \cite{sebanz2009prediction}. For example, the ability to predict temporal stimuli in a non-social context influences the ability to synchronize actions with a human partner \cite{pecenka2011role}, indicating that prediction in time plays an important role. When following the actions of a brick stacking human, observers' eye movements are usually anticipative instead of reactive \cite{flanagan2003action}, indicating prediction in space. Finally, observers expect others' to act efficiently towards their goals and in accordance with the current environment and situation \cite{koster2013theory}, indicating the prediction of actions in context over several seconds.   

% Prediction and robots
\paragraph{A robotics viewpoint}
 
In collaborative HRC, the prediction of the human's movements is of high importance. First of all, the robot should not plan a movement that might interfere with the human's limb positions in order to guarantee safety and comfortable interactions. For example, a hand-over task requires the robot to be able to predict the location of the hand-over. Secondly, anticipation of movements allows a robot to initiate actions in time such that the human does not have to wait for completed actions. Thus, when a human initiates a hand-over action, the robot should initiate the taking-action before the human has reached the intended hand-over location. 
 We elaborate on the topic of prediction and anticipation in HRC in Section \ref{secsec:prediction_robot}.

%----------------------------------------------------------------------------
% signaling in humans 
\subsubsection{Signaling} 
\label{sec:signalinghum}

As discussed above, prediction of others' actions plays a fundamental role in social interaction. Within the context of a shared task, the acting partner can enhance his or her predictability by actively signaling  intentional and attentional states. 
These signals can be very subtle such as changing the kinematics of the movement trajectory. 
% kinematic signaling
This includes adjusting the trajectory and the velocity and acceleration profile of a movement and reducing intertrial variability of movements \cite{sacheli2013kinematics, sartori2009does, scorolli2014give}.  However, as these adjustments deviate from the energy efficient default kinematics, these changes need to balance the cost and benefit of signaling \cite{candidi2015interactional}.  

% gaze and head
More obvious signaling strategies are gaze and head direction and gestures and facial expressions. Gaze behavior has been found to be an important factor for turn-taking \cite{ho2015speaking}, joint attention \cite{frischen2007gaze}, intention inference \cite{huang2015using} and to coordinate actions \cite{vesper2016role}. Facial expressions can be seen as an expressway for emotional exchange which is modulated by the social situation \cite{jack14}.
% gesture
Gestures add to this repertoire by expressing mental imagery and spatial dimensions \cite{hostetter2008visible}.
 
% signaling and robots
\paragraph{A robotics viewpoint}

Forward models play a different role in robotics when a human is involved as the outcome of an action does not solely depend on the robot's action and state but also on human interference. Therefore, modeling the social signals send by the human is crucial for successful and fluent interaction. In order to make these inferences, the robot needs to be able to predict human movements, as discussed in Section \ref{secsec:prediction_robot}. Furthermore, the robot can enhance the performance and human perception of the interaction by actively signaling intend, which we present in Section \ref{sec:signalrobots}.

\section{HUMAN-ROBOT COLLABORATION}
\label{sec:humrob}

% -----------------------------------------------------------------------------
% Intro to HRC
% -----------------------------------------------------------------------------

In the previous section we presented essential mechanisms that underly the efficiency of HHI. Many of these findings have been considered by the HRC research community. In this section we are aiming at summarizing these approaches in the context of socSMCs.

Research in HRI includes problems that are concerned with how to "\textit{understand and shape the interactions between one or more humans and one or more robots}" \cite{goodrich2007human}. Different environments and tasks require the robot to exhibit different levels of autonomy, adaption, learning ability and compliance. In this work, we aim at elaborating problems that need to be solved to enable physical human-robot collaboration. We do not view the robot in a master-slave setting but want to emphasize how autonomous behavior in social interaction can be achieved with help of sensorimotor contingencies. The representation of the socSMCs in HRC is discussed in Section \ref{sec:smcsrob}.
In the previous sections we discussed how humans can engage in non-verbal communication and joint action by relying on a constant exchange of sensorimotor information. Using these findings as inspiration and guide lines, we will present a coherent picture of check SMCs and sync SMCs in HRC in the Section \ref{sec:checkrob} and \ref{sec:synccouprob} respectively and point out how these mechanisms can give rise to active HRC in Section \ref{sec:jointactionrobot} and \ref{sec:syncplanrob}. Finally, we shortly introduce evaluation methods for performance assessment in HRC tasks in Section \ref{sec:evaluationrob}.

\subsection{\textbf{Embodied intelligence in a social context}}
\label{sec:smcsrob}
% -----------------------------------------------------------------------------
%  SMCs in robotics
% -----------------------------------------------------------------------------
The idea of embodied intelligence has influenced research in artificial intelligence and robotics since the 1990ies  \cite{pfeifer06}. The focus shifted from logic-based systems and static knowledge representations towards self-learning of action-perception-effect loops. Instead of a predefined high-level representation of the environment, embodied intelligence proposes that a representation solely based on sensory and motor signals suffices to result in intelligent behavior. For interaction with other agents, this approach brings several advantages. Firstly, many already acquired action-perception loops can be utilized for action recognition and prediction of a human partner. Secondly, as the human's actions are directly encoded in the robot's sensory state, it becomes natural to design safe and intuitive systems. Thirdly, decision making and planning under uncertainty is facilitated since the sensorimotor representation confines the search space of possible actions. In Section \ref{sec:repsmcsrob} we elaborate on the sensory channels that are commonly used in this research area.

% -----------------------------------------------------------------------------
\subsubsection{ SMCs - Representation in sensorimotor spaces} 
\label{sec:repsmcsrob}
 % work from past project? Pure smcs, object oriented smcs
In the human body, sensory and motor signals in nerves and muscles are interdependent and have a common coding space - the brain. Thus, sensorimotor contingencies are learned effortlessly during childhood and can even be acquired in adults (see e.g. \cite{kaspar14}). In order to provide the same level of instantaneous signal integration in robotic agents, sensory and motor signals can be combined on different time and precision scales. The main objective is to design systems that require a minimal amount of assumptions and that can learn to interact with the environment autonomously.

% sensors and motors
For a given task, different sensors can be utilized. While a navigation task might require distance sensors \cite{MayeE13}, other systems have been equipped with tactile feedback \cite{Saal10} and visual signals \cite{montesano08}. Due to the advances in deep learning, even raw visual pixel input has been successfully integrated \cite{levine16, ghadirzadeh2017deep}. The state of the robot can be included in form of joint configurations, velocity and force/torque measurements \cite{ghadirzadeh16, wahlstrom15}, while action signals can range from low-level torque or velocity commands  \cite{levine16, ghadirzadeh16, ghadirzadeh16b} to high-level actions such as "pushing" \cite{montesano08, ivaldi14}.

% emergent behavior
Once the sensory input and motor output modalities have been established, the system is required to learn sensorimotor contingencies and to utilize them for meaningful behavior. 
% interacting with the environment

% -----------------------------------------------------------------------------
% check SMCs
% -----------------------------------------------------------------------------

\subsection{\textbf{check SMCs - a robotic observer}}
\label{sec:checkrob}

As described in Section \ref{sec:checkhuman}, check SMCs include a sensorimotor coupling from one agent to another agent. In the following sections, we will elaborate on what this implies in terms of HRC. While humans might show signs of coupling with respect to a robot, we will focus on sensorimotor adaption to a human. We begin by discussing two fundamental problems, the matter of representing a human (Section \ref{sec:representationrobot}) and the mapping between embodiments (Section \ref{sec:corrprob}). Further, we present imitation learning, both in the context of  movements and affordances in Section \ref{sec:affordance}. Imitation does imply that the robot adapts its motor behavior to the perceptual input it receives from the human, i.e. it belongs to the category of check SMCs.   

% -----------------------------------------------------------------------------
\subsubsection{Representation of the human partner}
\label{sec:representationrobot}

The human partner can be incorporated into decision making indirectly, e.g. through force signals \cite{ghadirzadeh16b}, or explicitly modeled from visual input. 
While a pure sensorimotor approach to cognition is reluctant to model internal representations of the external world, for many computational models  a certain level of assumptions about representations is required. Thus, for human action recognition and understanding, a model of the human body in form of joints or silhouettes can be used to map changes of the body posture to actions  (see \cite{weinland11} for an in-depth discussion of vision-based action representations). Other approaches in computer vision use hand-crafted or learned image features to represent action induced changes in images. Such methods are highly perception-based with only a minimal amount of assumptions. However, these methods are less prevalent in HRC as the mapping between embodiments is impeded and requires a skeletal model. This mapping between embodiments is discussed below.

% -----------------------------------------------------------------------------
\subsubsection{Learning a mapping - the correspondence problem}
\label{sec:corrprob}
% Learning a mapping - the correspondence problem
A fundamental problem of embodied HRC is to establish a mapping between an observed human body and the robot's own embodiment \cite{nehaniv02}. In order to imitate human actions, the robot needs to acquire a model or function of how the human motion can be translated into its own motor commands.  The complexity of this function depends on different factors. First of all, the  difference of degrees of freedom between human and robotic actuators influences whether an analytic solution, i.e. a one-to-one mapping, is feasible \cite{mohammad13}. More often however, machine learning techniques are applied instead of analytic solutions which require the ability to generalize well from the training data to unseen configurations. For this, probabilistic methods such as Gaussian Mixture Models \cite{calinon07} and Gaussian Process Latent variable models \cite{shon05, lei15} have been used. The same concepts do not only apply to imitation of body movements but also facial expressions \cite{vitale14}. Next to the degrees of freedom, the sensory channels of the robot with which the human actions are sampled is vital. As pointed out by \cite{zuher12}, most developed techniques for learning by demonstration suffer from the need of advanced human motion capture techniques which are not available in most natural settings. For the purpose of HRC, purely vision based systems are of high importance \cite{lei15}.  Finally, the level of the mapping is not restricted to motion trajectories but can build on task-level objectives \cite{schaal03}. This high-level imitation is discussed below.

% -----------------------------------------------------------------------------
\subsubsection{Affordances and imitation learning}
\label{sec:affordance}
% Affordances and imitation learning
% imitation
A robot can acquire a mapping from its own pose to the human body by randomly performing actions and being imitated by a human. In \cite{boucenna2014learning}, the correlations between these two entities are modeled with neural networks. The authors find that imitating adults is less demanding than imitating children. Additionally, children diagnosed with autism spectrum disorder are more challenging than typically developing children. Thus, the imitation of human poses does not only depend on the mapping as discussed above in Section \ref{sec:corrprob} but also on the structure of the demonstration.

In comparison to correlation based approaches as described above, a different approach is to make use of dynamical movement primitives (DMP). DMPs cast the problem of control into a framework of dynamical systems  combined with non-parametric regression. They can be used to either learn behavior from human demonstration \cite{ijspeert2003learning} or to acquire interactive movement primitives in a joint task \cite{amor2014interaction}.     

% affordances
High-level imitation is often associated with exploiting affordances. Affordances belong to the key concepts of embodied cognition \cite{gibson14}. Instead of encoding objects by their type, e.g. a chair is a chair, they are encoded by associations with those actions they afford. A chair affords to sit on, to be pushed under the table, to climb onto for reaching the top of the shelf, etc. Affordance based behavior follows the action opportunities that the environment offers. This concept has been adopted by the area of robotics as knowledge and world representation are reduced to interaction and exploration of the environment \cite{horton12}. Several approaches towards affordance learning have involved active exploration of action-perception-effect contingencies, mostly in the visual domain \cite{montesano08, sun10} but also haptic feedback has been considered \cite{chu16}. As affordances are a generally applicable concept, humans and robots share most affordances. Thus, once learned, these properties can be used to imitate human actions or to predict future actions \cite{koppula16}. On the other hand, object affordances can be acquired by observing manipulations by a human \cite{kjellstrom11, koppula13}.

In HRC, espescially during physical interaction, a common understanding of the shared workspace is needed. The concept of affordances offers a tool to establish this understanding and enable human-robot teams to work towards a shared goal. Imitation in this context is more than learning by demonstration but a natural way of acquiring new SMCs. Not only object oriented actions need to be learned but also socially important actions and their interpretations, such as gaze and facial expressions or hand-shaking need to be understood. However, these signals depend on both the sender and the receiver, i.e. are to be categorized as sync SMCs which are discussed below.

% gestures and social affordances 

% -----------------------------------------------------------------------------
% sync SMCs mutual coupling
% -----------------------------------------------------------------------------
\subsection{\textbf{sync SMCs - mutual coupling}}
\label{sec:synccouprob}

As the transition from check SMCs to sync SMCs is smooth, a clear distinction of the two is not possible. In robotics, the problem of sensorimotor coupling such as synchronization and turn-taking is often studied from a single-agent perspective. This means that either the entrainment of the human towards the robot is studied or techniques to produce similar behavior in robots are investigated. However, mutual entrainment, i.e. active human-robot coupling, as it would be demanded by the definition of sync SMCs, is rarely addressed. Nevertheless, we review here work that is concerned with coupling in human-robot duos even if only one partner is actively adjusting to the other's actions. 

We begin by discussing synchronization and entrainment in human-robot teams in Section \ref{sec:synchronizationrobot}. Similar mechanisms are responsible for turn-taking, presented in Section \ref{sec:turntakingrobot}, and joint attention, presented in Section \ref{sec:jointattentionrob}.

% -----------------------------------------------------------------------------
\subsubsection{Synchronization and entrainment}
\label{sec:synchronizationrobot}
 
% synchrony
Entrainment between humans and robots does highly depend on a continuous exchange of sensorimotor signals. 
It can be both unidirectional, i.e. only one partner adapts, and mutual, i.e. both partners change their behavior. If only one partner adjusts actions towards the others' behavior synchronization can arise but might be less stable. As shown in  \cite{lorenz11}, humans tend to synchronize with both human and robotic partners. In comparison to the HHI setting however, the synchronization pattern in the HRC setting is less stable as the robotic movements were constant and non-adaptive. This prohibited a mutual coupling between the partners. 

In order to overcome this lack of reciprocity, M{\"o}rtl et al. \cite{mortl14} apply a continuous dynamical process  to a repetitive joint task with multiple movement primitives. These potential events for synchronization are successfully used to generate a higher level of synchrony. As in the case of entrainment in HHI tasks, subjects rated trials with a high synchrony level as more pleasant. The idea of neural oscillators for learning and adjusting rhythmic movements has previously e.g. been applied to learning drumming and to drawing a figure-8 \cite{ijspeert02}.

Multi-modal input modalities can assist the robot in learning to react in accordance with human actions. In  \cite{crick2006} a drumming scenario is devised in which the robot receives visual, auditory and proprioceptive signals. These multi-modal signals are used to adjust internal oscillators to achieve synchronous drumming even under noisy conditions. 

Instead of solely aiming at creating synchronous behavior, in \cite{prepin2010agent} the level of synchrony is used as a reinforcment signal for relevant information extraction and action adaptation. The synchrony level influences the learning of associations between perceptual input and motor output, enabling autonomous learning of socSMCs.

In \cite{miyake2009interpersonal}, the usefulness of interpersonal entrainment in medical applications is demonstrated. Employing a hierarchical nonlinear oscillator system, the authors show how the gait of Parkinson's patients stabilizes when they receive auditory feedback of a virtual walking partner.

% -----------------------------------------------------------------------------
\subsubsection{Turn-taking} 
\label{sec:turntakingrobot}

Next to synchronization and entrainment reciprocity can result in high-level coordination of behavior such as turn-taking. Turn-taking requires to read social signals and active engagement of all partners. Dautenhahn et al. \cite{broz12} gave a robot positive and negative feedback for socially engaging and turn-taking behavior and overlapping or non-engaged behavior respectively in an interactive game. With help of a short-term memory of the recent interaction history, the system learned  appropriate turn-taking actions.

In \cite{leite13} it was shown that simple machine learning algorithms to acquire turn-taking rules result in more human-like behavior. Nevertheless, the majority of studies on turn-taking in HRC relies on predefined rules found in HHI \cite{calisgan12, hart14}. As this approach can result in stereotypic behavior, a good understanding of the interaction patterns and a continuous interpretation of the interaction are required to find relevant cues and react more intuitively for the human partner \cite{skantze14}.

% ----------------------------------------------------------------------------- 
\subsubsection{Joint attention} 
\label{sec:jointattentionrob}

Joint attention is the state of two or more agents that share a common objective of focus, e.g. a certain object in the environment.
Both the concepts of synchronization and turn-taking are important to establish joint attention. Mutual guidance of attention to salient cues and constant reciprocal signaling is needed in order to maintain a common center of focus. For HRC this implies that not only the a attention following mechanism for the robot is needed but that a mutual understanding needs to be brought about.   Gaze is a salient cue to direct attention towards objects of interest in a shared task setting.
While infants can predict referential objects from human gaze signals at an age of 12-months they fail to make similar inferences for robotic gaze \cite{okumura13}. This suggests that humans need to learn to adapt to artificial cues. Similarily, \cite{admoni2011robot} showed that the automaticity with which human gaze is interpreted does not seem to hold for robotic agents. 

In order to make gaze comprehensible for the human, the two partners need to share a 3D environment \cite{moubayed2013}.  Furthermore, the frequency and duration of the gaze impact the signal strength of the robot's gaze \cite{admoni2013you}.

To learn the correlations between gaze direction and the environment, the robot is assisted with pointing gestures in \cite{doniec06}. In a different approach, the human partner evaluates the level of joint attention such as to teach the robot to follow gaze cues \cite{nagai06}. Furthermore, saliency cues such as which objects the human interacts with and task-related features can contribute to decrease the size of the search space for salient areas \cite{yucel13}.

% -----------------------------------------------------------------------------
% sync SMCs engaging in joint action
% -----------------------------------------------------------------------------
\subsection{\textbf{sync SMCs - engaging in joint action}}
\label{sec:jointactionrobot}

The basic principles governing check SMCs and sync SMCs that have been discussed up to this point are building blocks for higher-order coordination and active collaboration towards a shared goal. This aspect of sync SMCs goes beyond the coupling of low-level sensorimotor signals and takes into account active planning and role-sharing. In this section, we discuss how the historically prominent research directions of the master-slave scheme and imitation learning can be overcome when the problem of collaboration is viewed from a sensorimotor perspective in Section \ref{sec:rolesrobots}. Specifically, we focus on the sensorimotor processes that are required to coordinate actions in space and time in order to achieve a shared representation of the environment, such as mutual adaptation (Section \ref{sec:mutualadaprobot}), prediction (Section \ref{secsec:prediction_robot}) and signaling (Section \ref{sec:signalrobots}).

% -----------------------------------------------------------------------------
\subsubsection{Role division in joint action} % From slaves to equality
\label{sec:rolesrobots}

The master-slave principle casts interaction into an asymmetric relationship in which one partner takes a leading role while the other is a compliant follower \cite{jarrasse14}. This setup lends itself to scenarios in which a robot is treated as a physical assistant for task completion, such as lifting heavy objects. However, proactive behavior has been found to increase the efficiency of an interaction while decreasing the effort required from the human side \cite{lawitzky2010}. In order to allow an adjustable role distribution with smooth transitions, Kheddar et al. \cite{evrard09, kheddar11} introduced a homotopy that allows fluent changes between leader and follower roles. Depending on haptic and force measurements, the robotic system can adjust its behavior during the interaction. Hirche et al. \cite{mortl2012} studied how the manner of transitioning between behaviors influences the task performance and subjective experience of the human partner. In a table maneuvering task, the robot could either demonstrate a constant force, smoothly transition between minimum and maximum force as required by the situation or discretely jump between different force levels. While the continuous transitions were physically and mentally less demanding for the human subjects, they reported that they were under more control over the table in the discrete case. Additionally, the discrete change resulted in the most agreement between the partners. These results suggest that abrupt changes in behavior, although unpredictable and surprising in many cases, indicate clear transitions between roles and carry more information about the robot's state than the continuous case. In order to unravel the underlying dynamics  required for successful role-sharing with humans in physical HRC, more inputs from HHI might be required. Especially signaling strategies to avoid unpredictable behavioral changes need to be considered \cite{jarrasse14}.

% -----------------------------------------------------------------------------
\subsubsection{Mutual adaptation in joint action}
\label{sec:mutualadaprobot}

In addition to transitioning between roles, mutual adaptation is required to overcome the master-slave principle. Mutual adaption implies that both partners can infer each other's preferences and adapt their own behavior if necessary. Without an adapting robot, the team performance increases slower and to a lower degree than in a mutual adaptation setting \cite{ikemoto2012physical}. 

As different users might have different preferences in task execution, in \cite{nikolaidis15} these preference are modeled in order to adapt to a specific behavior type. The users are clustered according to their discrete action transition sequences and inverse reinforcement learning is applied to determine optimal behaviors for the robot. The results demonstrate faster task completion and more responsiveness towards the human as compared to a system designed by a domain-expert. Thus, robotic adaptation and human behavior modeling decreases the cognitive load of the human partner in a joint task. 

To model  human behavior, in \cite{nikolaidis2016formalizing} a bounded-memory adaptation model is introduced which relies on two assumptions. Firstly, it assumes that humans act based on a bounded history of states and secondly, that they adapt to a partner to varying degrees. In order to guide the human towards its own preferred goal, the robot includes inferences made with this model into its decision making and attempts to approximate the human's adaptability. If a human does not adapt, the robot initiates adaptation so as to maintain trust.

% Prediction in HRC -----------------------------------------------------------------------------------------------------------
\subsubsection{Predicting in joint action}
\label{secsec:prediction_robot}

Prediction in joint action occurs on three levels. When directly physically coupled to a human, the immediate sensory change due to human actions needs to be predicted. On the other hand, in an interaction, where exchange of physical signals is only temporary, the onset and offset of these periods need to be predicted. Finally, on a higher level, abstract actions towards a shared goal have to be inferred. 

% low level prediction  -  on sensory level
For low-level prediction, the incoming changes of the visual state caused by human movements have been modeled with e.g. Kalman filters. As soon as predictions can be made with high confidence, the robot can change from reactive to active behavior allowing for collaborative actions \cite{thobbi11}. In \cite{ghadirzadeh16b}, the change in interaction forces introduced by a human partner on a jointly held wooden plank is predicted by Gaussian Process forward models. 

% middle level prediction - when does sensory input suddenly change
Next to direct physical contact, the future trajectory of human limbs has to be estimated in order to avoid collisions and to prepare for future contact.  In \cite{kuderer12}, a trajectory is encoded by empirically defined motion features, such as velocity, acceleration and collision avoidance. By fitting a  probability distribution to these features, the system learns a human like navigation behavior for collision avoidance. Similarly, in \cite{mainprice2013human} human motion trajectories are modeled with help of Gaussian Mixture Models to estimate the space that will be occupied of the human.    

Instead of avoidance, the robot needs to plan motions towards a future contact point in handover tasks. In \cite{micelli2011perception}, the human's pose is used as an indicator for the intention to hand an object over. To account for the uncertainty in human behavior, dynamic movement primitives are adjusted in \cite{prada2014implementation} to account for the changing endpoint of the handover. 
%\textbf{Note: update references to CVPR when available} %%%%%%%%%%%%%%%%%%%%%%%%%%%%%%%%%%%%%%%%%%%%
While most approaches to endpoint prediction rely on task-specific hand trajectories, B\"utepage et al. \cite{butepage2017deep} introduce a deep learning approach to the problem of motion prediction. With help of encoding-decoding networks, they predict skeletal human pose data for up to 1600 ms based on a window of past observations. In a follow-up study, conditional variational autoencoders are applied to model a distribution over future observations \cite{butepage2017anticipating}. This structure allows both to sample possible future movements and to classify endpoints relying on motion predictions without task specific training data.

% high level prediction - action prediction
While predictions of low-level sensory and state changes are important, high-level action prediction is needed for collaborative task planning. As human actions are mostly intent driven, the recognition of intention and appropriate predictions of future behavior can simplify this complex problem  \cite{broz13}. An intention-driven model can facilitate anticipation of future actions and inferences over the internal state of an interaction partner. Highly dynamic interaction scenarios, such as playing table tennis \cite{wang13}, benefit from intention modeling. In \cite{wang13}, the intention of the human is defined as the outcome of an action and is modeled with a Intention-Driven Dynamics Model. This approach assumes intentions to be a latent variable that governs the latent dynamics and observed variables.  
To capture a more complicated cost function  inverse reinforcement learning is applied to  data from human-human interactions in \cite{mainprice15}. Subsequently, this cost function is used to iteratively predict human actions in a collaborative reaching task.

% Signaling in HRC ----------------------------------------------------------------------------------
\subsubsection{Signaling in joint action}
\label{sec:signalrobots}

Signaling in joint actions carries two different meanings. Firstly, it aims at direct signals that can guide the partner's attention, i.e. guide the high-level interaction. Secondly, subtle, low-level changes in kinematics and action profiles can serve as indicators for the intended goal, properties of the task or required effort. Since the guidance of attention is discussed in Sec. ~\ref{sec:jointattentionrob}, we will here mainly focus on the second part.

Goal-directed actions in a shared workspace have been characterized as a trade-off between legibility and predictability  by Dragan et al. \cite{dragan2013legibility}. On the one hand, a motion needs to be legible, i.e. an observer needs to be able to infer a goal given a trajectory. On the other hand, motions should not deviate too much from what is expected, i.e. they need to be predictable. In \cite{dragan2013legibility}, predictability of grasping gestures is defined as a function of action efficiency with respect to the goal.  While legibility is well-defined by the likelihood of predicting the correct goal given an observed trajectory, predictability depends on the task at hand and the objective view of the observer. This is supported by a study presented in \cite{dragan2015effects}, showing that humans tend to rate legible motion to be predictable in a social context in contrast to a neutral context. 

In a similar setting, the legibility of pointing gestures is studied in \cite{holladay2014legible}. Under the constraint of minimal deviation from a direct pointing gesture, it is shown that slight deviations for higher legibility increase the chance that human subjects infer the object of interest correctly.  

Not only goal-directed actions require signaling. For example, handover tasks might require an estimate of the object's weight to avoid surprise. To this end, weight aware lifting of objects is studied in 
\cite{sciutti2014understanding}, where it is shown that a velocity profile that matches the object's weight enables human observers to infer the weight with a higher accuracy. 

To signal the onset of an interaction, \cite{cakmak2011using} studies how contrasting social actions from other movements in terms of spatial and temporal variations can help humans to identify the intent of the action which results in smoother interactions.

% -----------------------------------------------------------------------------
% sync SMCs active collaboration
% -----------------------------------------------------------------------------
% Active collaboration

\subsection{\textbf{sync SMCs - active collaboration}}
\label{sec:syncplanrob}

The concepts introduced in the previous Section \ref{sec:jointactionrobot} such as mutual signaling and predictions are the foundation of active collaboration. As there exists no clear boundary between cooperation and collaboration in robotics research, we do not aim at selectively reviewing work concerned with collaboration as defined in Section \ref{sec:intro}. Instead, we want to focus on the concepts that are important to achieve active collaboration. We begin by discussing how to learn to collaborate, i.e. how to integrate socSMCs into the learning process in Section \ref{sec:learningcollaboraterobot}. As active decision making in a social context requires an understanding of one's partner's intentions and goals, we present work concerned with intention inference in Section \ref{sec:intentioninferencerobot}. Finally, collaboration requires long-term action planning that takes the actions of the human into account. Thus, we elaborate on collaborative task planning in Section \ref{sec:collaborativetaskplanningrobot}.

% learning in collaboration ----------------------------------------------------------------------------------
\subsubsection{Learning to collaborate}
\label{sec:learningcollaboraterobot}

Learning in interaction can be viewed from two positions. On the one hand, the learning itself can be a collaborative process in which one partner helps the other to learn a task \cite{kartoun2010human}. On the other hand, a robot can learn a joint task together with a human by incorporating human actions and task knowledge. We will here focus on the latter case.  

% actively asking the human for help
Robot learning in a joint setting can benefit from the human's task knowledge.
In order to assign semantic meanings to subphases of a coordination task, in \cite{medina2011experience} the robot  models SMC patterns in a positioning task with help of HMMs and asks the human to label consistent trajectory parts with semantic information. Through this, the robot benefits from the human task knowledge and can cluster trajectories according to this structure. 

A similar idea is presented in \cite{hayes2014discovering} which concentrates on active learning of a structured action-space  graph that describes different action-state sequences leading to the same goal. To gain more information about possible sequences, the robot can actively chose a skill that it requires additional information about. An informed selection, such as distance-based or connectivity-based, leads to the discovery of significantly more valid state-action sequences.

% multimodal is important
Next to explicit task knowledge, learning  robust representations and policies requires the robot to efficiently incorporate the multimodal sensorimotor data in a collaborative settings. 
Rozo et al. \cite{rozo2013learning} focus on a joint assembly task and investigate how to approximate stiffness parameters for different phases of the interaction. Applying GMMs to the task, the robot learns soft clusters of different behaviors which allows for reactive control. The authors stress the importance of combining visual and haptic feedback during collaboration as it allows to disambiguate different task states and human actions. % cooperation

Similarly, Huang et al. \cite{huang2016anticipatory} combine gaze tracking,  speech  recognition and intent  prediction to facilitate anticipatory  motion  planning, speech  synthesis and robotic manipulation. While the authors apply support vector machines to gaze patterns to classify intended objects, the failing rate of around 20 \% suggests that the inherently uncertain nature of human behavior might require probabilistic methods.  

A Bayesian approach to the problem of multimodal input data in physical HRC is presented in \cite{ghadirzadeh16b}. Leveraging Gaussian Processes to learn forward models and a Q-function, Ghadirzadeh et al. train a PR2 robot to position a ball on a wooden plank that is jointly controlled with a human partner. The sensory changes caused by human movements are encoded in force-torque measurements. Together with information about the robot's and game's state, these multimodal signals allow learning a reward function for optimal action selection in a collaborative task. 

In order to incorporate several high-level modalities such as speech and gestures, Huang et al. \cite{huang2014learning} differentiate between a feature level, describing the high-level behavioral features, and a domain level, describing behavioral features of specific modalities. Modeling speech, gestures and gaze of with Dynamic Bayesian Network, it is shown that the robot can learn communicative, multimodal behavior that is perceived more natural and effective than an unimodal equivalent.  

% multiple partners

% intention  inference ----------------------------------------------------------------------------------
\subsubsection{Intention inference}
\label{sec:intentioninferencerobot}

The term intention is associated with different meanings in different contexts. A general definition of intention might be the motivation or goal that let to the initiation of an action. Intentions can operate on different time scales and be hierarchical in nature. For example, imagine someone is reaching for the sugar in the kitchen. The short term intention is to acquire the sugar. This intention might be governed by a longer lasting intention to bake a pie. In turn, this intention might be influenced by the intention to give the pie to a friend as a social gesture. In terms of sensorimotor patterns, short term intentions are of relevance as they provide context for prediction of actions and states and because they allow for assisting, complementary and avoiding action selection in a collaborative setting. 

In \cite{li2014human}, the human motion intention is defined as the resting state of a mass-spring system that describes the dynamics of the human arm. Others define intention as the final outcome of an action, such as the goal position of an object in the workspace \cite{wang13,whitney2016interpreting}.
In order to infer intentions, one can either rely on state information alone, such as the current configuration of the human joints, or on a combination of states and actions. In the following, we will elaborate on techniques used for human intention inference that are based on sensory and state and action signals. 

% state based
Intention on a motor level as defined in \cite{li2014human}, can be inferred based on current state, velocity and interaction forces. In \cite{li2014human}, a radial basis function neural network is used to learn a function that maps the current information to the intended outcome. Using these approximations, a robot is shown to follow human movements. The results indicate that less effort is required from the human in comparison to a common impedance controller in this leader-follower task.  
In direct cooperation settings, a mixture of interaction primitives can be used to model human movement trajectories with different intentions \cite{ewerton2015learning}. After classifying the current human primitive, the robot can engage in the interaction by following the learned interaction primitives. 

In a Bayesian view, the intention underlying a movement can be seen as a latent variable that can be modeled in a joint distribution with the observations. If high-dimensional observations are governed by low-dimensional dynamics, an Intention Driven Dynamics Model can be learned \cite{wang13} that models the joint distribution of observations, dynamics and intentions. During online interaction, the robot can infer the dynamics and intentions of the human and plan actions accordingly. 
A similar idea is proposed in \cite{whitney2016interpreting} who used a Bayesian filter to integrate multimodal signals. In a joint cooking scenario, the robot has to infer which objects the human might require based on the recipe, speech and gestures. The Bayesian filter allows to model the joint probability distribution of these variables and to effectively infer the identities of the objects. 

% state and action based
An interactive setting has the advantage that both agents can not only select goal-direct actions but also actions that force their partner to reveal hidden information. In \cite{sadigh2016information} a shared control problem is formulated that allows the machine to select actions which will push the human towards actions which express an internal state or preferences.

% Collaborative task planning ----------------------------------------------------------------------------------
\subsubsection{Action selection in collaborative tasks}
\label{sec:collaborativetaskplanningrobot}

We can distinguish between two approaches to action selection, reinforcement learning and classical planning. In reinforcement learning, the robot learns a value function and a policy over time, while planning relies on explicit lookaheads in the action-state space. When considered  in collaborative robotics, both methods builds on three components \cite{koppula2016anticipatory}. First of all, the environment and the context need to be modeled, which includes spatial references and affordances. Secondly, human actions need to be understood and anticipated in this context. Finally, the robot needs to choose actions and movements taking these uncertain estimates into account and considering how to facilitate human perception of the task.

% planning
Mainprice et al. \cite{mainprice2013human} consider a scenario in which a human and a robot operate in the same workspace. This is achieved with a trajectory planning algorithm, STOMP, that iteratively updates an initial trajectory. The authors show that modeling and incorporating the space that will be occupied by the human in the future compared to the current position results in faster convergence of the planning algorithm. Next to distance other interaction constraints can be incorporated into planning such as comfort, visibility and fluency constraints \cite{mainprice2010planning, mainprice2012sharing}.  Note, that these planning approaches view the human as a part of the environment that needs to be planned around instead of considering the human a collaborative partner.

In direct physical contact, human and robot need to negotiate the amount of effort each partner has to apply to achieve the common goal. Posing a shared positing task as a control problem with two forces that act on the object of interest, in \cite{mortl2012} the role division between the partners is investigated as discussed in Section \ref{sec:rolesrobots}. The authors distinguish between three policies: a balanced-effort policy, a maximum-robot-effort policy and a minimum-robot-effort policy. In an online collaborative task, the system needs to be able to 
switch between these policies based on the current state and the human behavior. 

%\begin{figure*}[th!]
%\centering
%\includegraphics [ width= 1 \textwidth]{figures/evaluation.png}
%\caption{To evaluate HRC and HRC, most studies follow either Scenario 1 (left) or Scenario 2 (right). The two rows indicate different time steps. In Scenario 1 the state representation is kept stable, while different actions are selected. The effects of different actions, i.e. approach a vs approach b, are compared in qualitative and quantitative measures.
%In Scenario 2 the state representation is varied, while the action selection process is kept stable. The effects of different states, i.e. approach a vs approach b, are compared in qualitative and quantitative measures.} \label{Fig:evaluation}
%\end{figure*}

In comparison to many studies that focus on how to select actions in the presence of an active human partner, the notion of legibility and predictability as discussed in Section \ref{sec:signalrobots} requires the robot to plan in the presence of a human observer in order to signal its intention. By formalizing this problem in mathematical terms, Dragan et al. \cite{dragan14} apply functional gradient and constraint optimization methods to generate legible and predictable trajectories. 

% reinforcement learning
While planning under interaction constraints results in actions which implicitly take the human into account, learning collaborative value functions has the advantage of explicitly incorporating human actions.

In \cite{koppula2016anticipatory}, collaborative tasks are tackled with multi-agent Q-learning. By representing states with help of affordances, a shared representation of the environment arises naturally. Comparing different models of human behavior, the authors conclude that a model of adaptive behavior outperforms static approaches. 

Instead of a multi-agent value function, Nikolaidis et al. \cite{nikolaidis2013human} show that cross-training of a human-robot team is more effective than human feedback on task performance. This cross-training implies that human and robot iteratively switch between the respective roles. Thus, they learn a shared plan instead of separate action sequences.

% -----------------------------------------------------------------------------
% Evaluation of embodied collaborative robotics
% -----------------------------------------------------------------------------
\subsection{\textbf{Evaluation of embodied collaborative robotics}}
\label{sec:evaluationrob}

% offline vs online
A systematic, general approach to evaluation in HRC is near to impossible as the field is so versatile. Within each subfield, however, common metrics have been established. One important aspect to consider is online vs. offline evaluation. While safety measures should be applied constantly, more qualitative measures such as the human perception of the interaction are mostly taken into account after the interaction has terminated.

% level of autonomy
An additional factor is the level on autonomy. While some studies work with a wizard-of-oz setup, i.e. the robot is controlled by a human, others focus on active decision making. Although these two approaches might apply similar evaluation methods, the results are not necessarily comparable. 

% test what additional SMCs bring to the interaction -> must be careful because you might just add more information!
Testing a scientific hypothesis with quantitative measures of SMCs does often imply to investigate the necessity of specific SMCs or behaviors compared to alternative approaches. We can either vary the reaction to a sensory state, or vary the sensory state itself. 
For example, in \cite{mutlu2013coordination} coordination mechanisms between humans and robots are investigated. Implementing a variety of robot behaviors to cue an intended object with gaze, the human's performance to detect this object are studied. 
In comparison, in \cite{whitney2016interpreting}, the input state to the robot is varied between receiving only gesture and both gesture and speech input. The effect of these variations are evaluated in the context of the interaction with quantitative and qualitative measures. As described in Section \ref{sec:quanmeascoll} and \ref{sec:qualmeascoll}, these measures can be either objective or rely on the subjective experience of the human.

% -----------------------------------------------------------------------------
% Quantitative measures of collaboration
\subsubsection{Objective measures of collaboration}
\label{sec:quanmeascoll}

There exists a variety of objective measures for human-robot collaboration. In this section, we will name the most relevant to the topic of embodied collaborative robotics. For a more detailed discussion, we refer the reader to reviews such as \cite{young2011evaluating}. 

% safe
\noindent \textbf{Safety}
The most important factor in HRC is that the human can safely interact with the robot. To guarantee this, the robotic system needs to be robust, context aware and reactive in dangerous situations, see e.g. \cite{giuliani2010design}.

% fluent
\noindent \textbf{Fluency}
Fluency of an interaction can be measured by how much each partner has to wait for the other partner to finish some action before it can act itself. This can be quantified by e.g. the amount of idle time for the human or robot, concurrent motion and the time between alternating actions, see e.g.\cite{hoffman2007effects}.

% engagement
\noindent \textbf{Engagement}
Engagement is a measure of how engaged the human is in the interaction with the robot, i.e. how much prosocial behavior is present and how distracted the subject is over the course of the interaction. This engagement might vary with the structure of the interaction, e.g. individuals compared to groups \cite{leite2015comparing} .

 % mutual information
\noindent \textbf{Mutual information}
A measure of the shared information content two sensory sources carry is mutual information. On the one hand, this measure can be applied to measure synchrony in received multimodal signals, see e.g. \cite{rolf2009attention}. On the other hand, it can give a measure of the reciprocity between two partners, see e.g. \cite{meisner2009shadowplay}. 

% completion time 
\noindent \textbf{Completion time}
Most interaction scenarios treated in HRC consist of short-term interactions with a specific goal. When this goal is reached, the trial is terminated. The statistical differences between conditions, when e.g. actions or states are varied, is interpreted as a measure of efficiency. 

% reliability of completion
\noindent \textbf{Failure rate}
Failure of a collaborative task might depend on the performance of both partners. Depending on the task at hand, the human might fail to interpret the signals sent by the robot or the robot might fail to perceive or understand the human or generate an appropriate action. The failure rate is the percentage of trials that were not completed satisfactorily. Instead of failing a task, a partner can also fail to cooperate, when viewed in a game theoretic context \cite{sandoval2016reciprocity}.

% resource utilization and cost
\noindent \textbf{Physical cost}
Physical cost is usually measured in terms of forces that both the human and the robot need to apply to achieve a task \cite{mortl2012}. Depending on the task, these forces should either be equally distributed, i.e. equal roles, or in favor of the human partner, i.e. a master-slave setting. 

% mental workload
\noindent \textbf{Cognitive load}
The cognitive load describes the amount of cognitive work and decision making the human is responsible in a joint task. A master-slave setting would encourage a high amount of mental workload, while an autonomous robot suggests an equal distribution. To estimate mental workload, the spare mental capacity of a human subject can be measured, i.e. the subject has to solve a secondary task next to the interaction with the robot. Additionally, physiological responses, such as heart rate or electroencephalography, can be an indicator of mental workload \cite{harriott2015measuring}.

\begin{figure*}[b!]
\centering
\includegraphics [ width=1\textwidth]{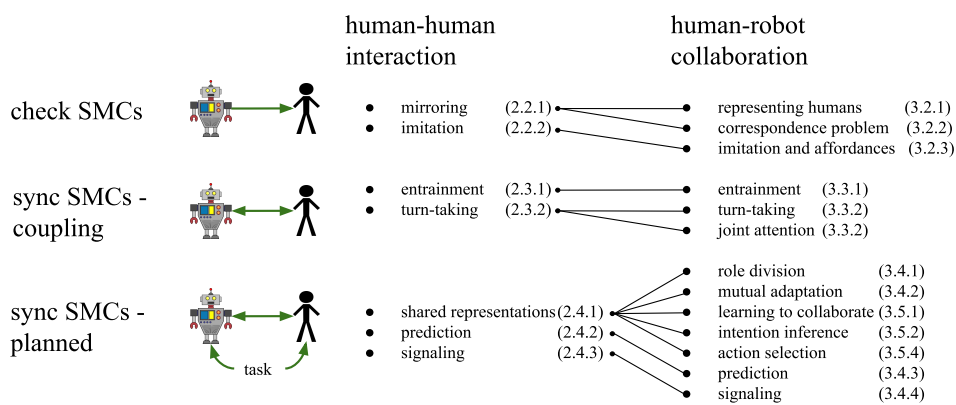}
\caption{The relationship between topics in HHI and HRC divided into check SMCs, sync SMCs mainly concerned with automatic entrainment and sync SMCs required for higher level coordination of actions. The numbers in brackets refer to the section of this review in which we discuss the respective topic. } \label{Fig:relations}
\end{figure*}

% -----------------------------------------------------------------------------
% Qualitative measures of collaboration
\subsubsection{Subjective measures of collaboration}
\label{sec:qualmeascoll}

Subjective measures of human-robot collaboration focus on the subjective experience of users. Thus, they might be influenced by the background and the training of the human subjects. 

% questionnaire 
\noindent \textbf{Questionnaire}

Trials with similar performance as measured by the quantitative measures described in Section \ref{sec:quanmeascoll} might be perceived very differently by human subjects. In order to quantify these differences, a common approach are questionnaires which measures aspects such as perceived safety or collaboration quality with likert scales, see e.g. \cite{bartneck2009measurement}.

% task plan 
\noindent \textbf{Shared mental plan }
In order to evaluate whether a human-robot team has achieved a shared mental plan of a structured task, the human subjects can explicitly indicate their planned action sequence and the agreement with the robot's plan can be estimated , see e.g. \cite{nikolaidis2013human}.

\section{CONCLUSIONS}
\label{sec:future}

Human-robot collaboration is a challenging subject as the presence of the human introduces uncertainty and constraints on the robotic system which are often not foreseeable during the system design phase. In the introduction we hypothesized that an embodied approach can facilitate the collaboration as it offers a basis for grounding the interaction. We enlisted three questions which have to be considered when aiming at autonomous robots in collaborative settings. In this section, we want to investigate to which degree these questions have been addressed by the community.

% question 1 ---------------------------------------------------------------------------------------
\subsection{\textbf{Which principles and mechanisms govern HHI that can inspire the design of HRC systems}}

We address this question in Section \ref{sec:humhum} by reviewing relevant literature concerned with HHI that can inspire research in HRC. There exists a vast body of research in this area of which we selected the most relevant aspects to HRC. We depict the influence of HHI on HRC in \textbf{Figure \ref{Fig:relations}}, where we distinguish between check SMCs, sync SMCs concerned with coupling of SMC processes, and sync SMCs that are involved in planning and performing a shared task.

A reoccurring structure in HHI is the fact that humans ground interaction in SMCs such that they share a representational basis. Within check SMCs, this is governed by the concept of mirroring others' actions, i.e. the existence of a SMC mapping between humans. This allows us to imitate and mimic others.  The coupling inherent to sync SMCs gives rise to automatic entrainment and turn-taking behavior. Finally, in order to reach shared goals, humans rely on shared representations. To avoid conflicts, the actions of a partner need to be predicted and one's own intentions have to be signaled.  

As depicted in \textbf{Figure \ref{Fig:relations}}, these concepts have influenced different areas of interest in HRC. 

% question 2 ---------------------------------------------------------------------------------------
\subsection{\textbf{What kind of state representations and mechanisms are optimal for HRC? }}

In an interactive setting, an egocentric world representation is not sufficient to allow collaborative behavior. Instead, the human needs to be modeled and included into decision making. Therefore, concepts originating from SMCs are of high value as they are shared with the human partner. For example, the affordances of an object can often be shared between the human and the robot, leading to a natural task representation. We discuss these representations mainly in the context of check SMCs in Section \ref{sec:checkrob} but they are also required for the emergence of sync SMCs as discussed in Section \ref{sec:synccouprob}.

The three main mechanisms driving HHI are a shared representation, prediction and signaling. These mechanisms guide interaction patterns such as role division, intention inference and coordination of actions. However, in the robotics community, the emphasis is not on the mechanism, i.e. a shared representation, but on the phenomena, see Section \ref{sec:jointactionrobot} and \ref{sec:syncplanrob}.

% question 3 --------------------------------------------------------------------------------------- 
\subsection{\textbf{How can we achieve autonomous, collaborative behavior?}}

Autonomous, collaborative behavior requires the interplay between the different mechanisms which have been presented in this survey. To achieve truly collaborative instead of instructive or cooperative interactions, it is important that the robot behaves naturally for the human such that mutual adaptation and learning can occur to guide the human-robot team towards shared goals. Since the large majority of studies in HRC focuses either on the robotic implementation or the human experience of the interaction, we are not aware of studies that have attempted to combine a shared representation, prediction and signaling to achieve true collaboration. 

Additionally, many tasks and scenarios in HRC might not require collaboration but only cooperation. However, humans tend to solve tasks collaboratively even when this is not required by the task \cite{warneken2012collaborative}. This indicates that collaboration is driven by an instinct to work together, deepening social bonding and opening the possibility to learn from each other. To achieve a similar behavior in robots, we are required to combine the mechanisms of interaction with the drive to engage with humans. Instead of only modeling the task performance, the robot can be rewarded by how well it predicts its human partner and how well the human responds to its own actions.  
Thus, although most scenarios do not require collaboration, they do allow it.

\subsection{\textbf{Future directions}}

To enable physical human-robot collaboration, a unification of the mechanisms introduced in this review is required. Sensorimotor contingencies are well suited as a representational basis since they allow a common coding of the robot's actions and the perception of human actions. To plan actions in a collaborative setting requires the robot to adapt to the human but also to signal when the human partner should change current behavior. In order to evaluate and compare different approaches, scenarios of the different levels, instruction, cooperation and collaboration need to be devised, which are easily reproducible by other groups.

\endgroup

%Disclosure
\section*{DISCLOSURE STATEMENT}
The authors are not aware of any affiliations, memberships, funding, or financial holdings that
might be perceived as affecting the objectivity of this review. 

% Acknowledgements
\section*{ACKNOWLEDGMENTS}
This work was supported by the EU through the project socSMCs (H2020-FETPROACT-2014) and Swedish Foundation for Strategic Research. 

% References
%
% Margin notes within bibliography

%\input{includes/bibli}
\bibliographystyle{ar-style3}
\bibliography{bib}

\begin{thebibliography}{181}
\expandafter\ifx\csname natexlab\endcsname\relax\def\natexlab#1{#1}\fi

\bibitem{jarrasse14}
Jarrass{\'e} N, Sanguineti V, Burdet E. 2014.
Slaves no longer: review on role assignment for human--robot joint motor
  action.
\textit{Adaptive Behavior} 22:70--82

\bibitem{argall09}
Argall BD, Chernova S, Veloso M, Browning B. 2009.
A survey of robot learning from demonstration.
\textit{Robotics and autonomous systems} 57:469--483

\bibitem{keast2007getting}
Keast R, Brown K, Mandell M. 2007.
Getting the right mix: Unpacking integration meanings and strategies.
\textit{International Public Management Journal} 10:9--33

\bibitem{engel2013s}
Engel AK, Maye A, Kurthen M, K{\"o}nig P. 2013.
Where's the action? the pragmatic turn in cognitive science.
\textit{Trends in cognitive sciences} 17:202--209

\bibitem{dipaolo12}
Di~Paolo EA, De~Jaegher H. 2012.
The interactive brain hypothesis.
\textit{Frontiers in Human Neuroscience} 6:163

\bibitem{butepage2016social}
B{\"u}tepage J, Kjellstr{\"o}m H, Kragic D. 2016.
Social affordance tracking over time-a sensorimotor account of false-belief
  tasks. In \textit{Proc. 38th Annual Meeting of the Cognitive Science Society
  (CogSci)}. Cognitive Science Society

\bibitem{brooks91}
Brooks R. 1991.
New approaches to robotics.
\textit{Science} 253:1227--1232

\bibitem{oregan01}
O'Regan JK, No{\"e} A. 2001.
A sensorimotor account of vision and visual consciousness.
\textit{Behavioral and brain sciences} 24:939--973

\bibitem{pfeifer06}
Pfeifer R, Bongard J. 2006.
How the body shapes the way we think: a new view of intelligence.
Bradford Books

\bibitem{wolpert07}
Wolpert DM. 2007.
Probabilistic models in human sensorimotor control.
\textit{Human movement science} 26:511--524

\bibitem{orban11}
Orb{\'a}n G, Wolpert DM. 2011.
Representations of uncertainty in sensorimotor control.
\textit{Current opinion in neurobiology} 21:629--635

\bibitem{haruno01}
Haruno M, Wolpert DH, Kawato M. 2001.
Mosaic model for sensorimotor learning and control.
\textit{Neural computation} 13:2201--2220

\bibitem{hogman13}
Hogman V, Bjorkman M, Kragic D. 2013.
Interactive object classification using sensorimotor contingencies. In
  \textit{Intelligent Robots and Systems (IROS), 2013 IEEE/RSJ International
  Conference on}. IEEE

\bibitem{Stork15}
Stork JA, Ek CH, Bekiroglu Y, Kragic D. 2015.
Learning predictive state representation for in-hand manipulation. In
  \textit{ICRA}. IEEE

\bibitem{boots11}
Boots B, Siddiqi SM, Gordon GJ. 2011.
Closing the learning-planning loop with predictive state representations.
\textit{The International Journal of Robotics Research} 30:954--966

\bibitem{wolpert03}
Wolpert DM, Doya K, Kawato M. 2003.
A unifying computational framework for motor control and social interaction.
\textit{Philosophical Transactions of the Royal Society B: Biological Sciences}
  358:593--602

\bibitem{maris96}
Maris M, Boeckhorst R. 1996.
Exploiting physical constraints: heap formation through behavioral error in a
  group of robots. In \textit{Intelligent Robots and Systems' 96, IROS 96,
  Proceedings of the 1996 IEEE/RSJ International Conference on}, vol.~3. IEEE

\bibitem{montesano08}
Montesano L, Lopes M, Bernardino A, Santos-Victor J. 2008.
Learning object affordances: From sensory--motor coordination to imitation.
\textit{Robotics, IEEE Transactions on} 24:15--26

\bibitem{loehr201313}
Loehr JD, Sebanz N, Knoblich G. 2013.
13 joint action: From perception-action links to shared representations.
\textit{Action science: Foundations of an emerging discipline} :333

\bibitem{jokinen09}
Jokinen K, McTear M. 2009.
Spoken dialogue systems.
\textit{Synthesis Lectures on Human Language Technologies} 2:1--151

\bibitem{uthus13}
Uthus DC, Aha DW. 2013.
Multiparticipant chat analysis: A survey.
\textit{Artificial Intelligence} 199:106--121

\bibitem{turing50}
Turing AM. 1950.
Computing machinery and intelligence.
\textit{Mind} :433--460

\bibitem{mehrabian71}
Mehrabian A. 1971.
Silent messages.

\bibitem{dragan14}
Dragan A, Srinivasa S. 2014.
Integrating human observer inferences into robot motion planning.
\textit{Autonomous Robots} 37:351--368

\bibitem{vitale14}
Vitale J, Williams MA, Johnston B, Boccignone G. 2014.
Affective facial expression processing via simulation: A probabilistic model.
\textit{Biologically Inspired Cognitive Architectures} 10:30--41

\bibitem{Moon14}
Jung MA, Troniak D, Gleeson BT, Pan MKXJ, Zheng M, et~al. 2014.
Meet me where i'm gazing: how shared attention gaze affects human-robot
  handover timing. In \textit{{ACM/IEEE} International Conference on
  Human-Robot Interaction, HRI'14, Bielefeld, Germany, March 3-6, 2014}

\bibitem{mainprice15}
Mainprice J, Hayne R, Berenson D. 2015.
Predicting human reaching motion in collaborative tasks using inverse optimal
  control and iterative re-planning. In \textit{{IEEE} International Conference
  on Robotics and Automation, {ICRA} 2015, Seattle, WA, USA, 26-30 May, 2015}

\bibitem{socSMCproject}
 ????
Socializing sensorimotor contingencies - socsmcs, fet proactive horizon 2020
  funded research project. In \textit{H2020-FETPROACT-2014)}.
  \url{www.socsmcs.eu}

\bibitem{caronna08}
Caronna EB, Milunsky JM, Tager-Flusberg H. 2008.
Autism spectrum disorders: clinical and research frontiers.
\textit{Archives of Disease in Childhood} 93:518--523

\bibitem{sebanz2006joint}
Sebanz N, Bekkering H, Knoblich G. 2006.
Joint action: bodies and minds moving together.
\textit{Trends in cognitive sciences} 10:70--76

\bibitem{wahn2015multisensory}
Wahn B, Schwandt J, Kr{\"u}ger M, Crafa D, Nunnendorf V, K{\"o}nig P. 2015.
Multisensory teamwork: using a tactile or an auditory display to exchange gaze
  information improves performance in joint visual search.
\textit{Ergonomics} :1--15

\bibitem{ernst2002humans}
Ernst MO, Banks MS. 2002.
Humans integrate visual and haptic information in a statistically optimal
  fashion.
\textit{Nature} 415:429--433

\bibitem{bates2005external}
Bates AT, Patel TP, Liddle PF. 2005.
External behavior monitoring mirrors internal behavior monitoring:
  error-related negativity for observed errors.
\textit{Journal of Psychophysiology} 19:281--288

\bibitem{brown2012role}
Brown EC, Br{\"u}ne M. 2012.
The role of prediction in social neuroscience.
\textit{Frontiers in human neuroscience} 6:147

\bibitem{sebanz2009prediction}
Sebanz N, Knoblich G. 2009.
Prediction in joint action: What, when, and where.
\textit{Topics in Cognitive Science} 1:353--367

\bibitem{sacheli2013kinematics}
Sacheli LM, Tidoni E, Pavone EF, Aglioti SM, Candidi M. 2013.
Kinematics fingerprints of leader and follower role-taking during cooperative
  joint actions.
\textit{Experimental brain research} 226:473--486

\bibitem{Knoblich11}
Knoblich G, Butterfill S, Sebanz N. 2011.
Psychological research on joint action: theory and data.
No.~54 in The psychology of learning and theory: Advances in research and
  theory. San Diego: Elsevier,  59--101

\bibitem{richardson08}
Schmidt RC, Richardson MJ. 2008.
Dynamics of interpersonal coordination. In \textit{Coordination: Neural,
  Behavioral and Social Dynamics}, eds. A~Fuchs, VK~Jirsa, Understanding
  Complex Systems. Springer Berlin Heidelberg,  281--308

\bibitem{pezzulo13}
Pezzulo G, Candidi M, Dindo H, Barca L. 2013.
Action simulation in the human brain: twelve questions.
\textit{New Ideas in Psychology} 31:270--290

\bibitem{prinz1990common}
Prinz W. 1990.
A common coding approach to perception and action. In \textit{Relationships
  between perception and action}. Springer,  167--201

\bibitem{van2013action}
Van Der~Wel R, Sebanz N, Knoblich G. 2013.
Action perception from a common coding perspective.
\textit{People watching: Social, perceptual, and neurophysiological studies of
  body perception} :101--119

\bibitem{rizzolatti04}
Rizzolatti G, Craighero L. 2004.
The mirror-neuron system.
\textit{Annu. Rev. Neurosci.} 27:169--192

\bibitem{Sliwa745}
Sliwa J, Freiwald WA. 2017.
A dedicated network for social interaction processing in the primate brain.
\textit{Science} 356:745--749

\bibitem{oberman2007human}
Oberman LM, Pineda JA, Ramachandran VS. 2007.
The human mirror neuron system: a link between action observation and social
  skills.
\textit{Social cognitive and affective neuroscience} 2:62--66

\bibitem{keysers14}
Keysers C, Gazzola V. 2014.
Hebbian learning and predictive mirror neurons for actions, sensations and
  emotions.
\textit{Philosophical Transactions of the Royal Society B: Biological Sciences}
  369:20130175

\bibitem{cook2014mirror}
Cook R, Bird G, Catmur C, Press C, Heyes C. 2014.
Mirror neurons: from origin to function.
\textit{Behavioral and Brain Sciences} 37:177--192

\bibitem{wang2016robust}
Wang H, Oneata D, Verbeek J, Schmid C. 2016.
A robust and efficient video representation for action recognition.
\textit{International Journal of Computer Vision} 119:219--238

\bibitem{lopes2010abstraction}
Lopes M, Melo F, Montesano L, Santos-Victor J. 2010.
Abstraction levels for robotic imitation: Overview and computational
  approaches. In \textit{From Motor Learning to Interaction Learning in
  Robots}. Springer,  313--355

\bibitem{Platek2005}
Platek SM, Mohamed FB, Gallup GGJ. 2005.
Contagious yawning and the brain.
\textit{Cognitive Brain Research} 23:448--452

\bibitem{chartrand99}
Chartrand TL, Bargh JA. 1999.
The chameleon effect: the perception--behavior link and social interaction.
\textit{Journal of personality and social psychology} 76:893--910

\bibitem{jones09}
Jones SS. 2009.
The development of imitation in infancy.
\textit{Philosophical Transactions of the Royal Society of London B: Biological
  Sciences} 364:2325--2335

\bibitem{meltzoff1988infant}
Meltzoff AN. 1988.
Infant imitation after a 1-week delay: Long-term memory for novel acts and
  multiple stimuli.
\textit{Developmental psychology} 24:470

\bibitem{nielsen2006copying}
Nielsen M. 2006.
Copying actions and copying outcomes: social learning through the second year.
\textit{Developmental psychology} 42:555

\bibitem{gleibs2016group}
Gleibs IH, Wilson N, Reddy G, Catmur C. 2016.
Group dynamics in automatic imitation.
\textit{PloS one} 11:e0162880

\bibitem{over2013social}
Over H, Carpenter M. 2013.
The social side of imitation.
\textit{Child Development Perspectives} 7:6--11

\bibitem{over2013children}
Over H, Carpenter M, Spears R, Gattis M. 2013.
Children selectively trust individuals who have imitated them.
\textit{Social Development} 22:215--224

\bibitem{ingersoll08}
Ingersoll B. 2008.
The social role of imitation in autism: Implications for the treatment of
  imitation deficits.
\textit{Infants \& Young Children} 21:107--119

\bibitem{Zivotofsky12}
Zivotofsky AZ, Gruendlinger L, Hausdorff JM. 2012.
Modality-specific communication enabling gait synchronization during
  over-ground side-by-side walking.
\textit{Human Movement Science} 31:1268--1285

\bibitem{richardson07}
Richardson MJ, Marsh KL, Isenhower RW, Goodman JRL, Schmidt RC. 2007.
Rocking together: Dynamics of intentional and unintentional interpersonal
  coordination.
\textit{Human movement science} 26:867--891

\bibitem{sacks1974simplest}
Sacks H, Schegloff EA, Jefferson G. 1974.
A simplest systematics for the organization of turn-taking for conversation.
\textit{language} :696--735

\bibitem{duncan1972some}
Duncan S. 1972.
Some signals and rules for taking speaking turns in conversations.
\textit{Journal of personality and social psychology} 23:283

\bibitem{meyer2015journal}
Meyer M, Bekkering H, Haartsen R, Stapel J, Hunnius S. 2015.
Journal of experimental child psychology.
\textit{Journal of Experimental Child Psychology} 139:203--220

\bibitem{macrae08}
Macrae CN, Duffy OK, Miles LK, Lawrence J. 2008.
A case of hand waving: Action synchrony and person perception.
\textit{Cognition} 109:152--156

\bibitem{keller14}
Keller PE, Novembre G, Hove MJ. 2014.
Rhythm in joint action: psychological and neurophysiological mechanisms for
  real-time interpersonal coordination.
\textit{Philosophical Transactions of the Royal Society of London B: Biological
  Sciences} 369

\bibitem{van09}
Van~Baaren R, Janssen L, Chartrand TL, Dijksterhuis A. 2009.
Where is the love? the social aspects of mimicry.
\textit{Philosophical Transactions of the Royal Society B: Biological Sciences}
  364:2381--2389

\bibitem{marsh09}
Marsh KL, Richardson MJ, Schmidt RC. 2009.
Social connection through joint action and interpersonal coordination.
\textit{Topics in Cognitive Science} 1:320--339

\bibitem{wiltermuth09}
Wiltermuth SS, Heath C. 2009.
Synchrony and cooperation.
\textit{Psychological science} 20:1--5

\bibitem{sebanz03}
Sebanz N, Knoblich G, Prinz W. 2003.
Representing others' actions: just like one's own?
\textit{Cognition} 88:11--21

\bibitem{doerrfeld2012expecting}
Doerrfeld A, Sebanz N, Shiffrar M. 2012.
Expecting to lift a box together makes the load look lighter.
\textit{Psychological Research} 76:467--475

\bibitem{Richardson07a}
Richardson MJ, Marsh KL, Baron RM. 2007.
Judging and actualizing intrapersonal and interpersonal affordances.
\textit{Journal of Experimental Psychology: Human Perception and Performance}
  33:845--859

\bibitem{meagher2014costs}
Meagher BR, Marsh KL. 2014.
The costs of cooperation: Action-specific perception in the context of joint
  action.
\textit{Journal of experimental psychology: human perception and performance}
  40:429

\bibitem{van2008understanding}
van Schie HT, van Waterschoot BM, Bekkering H. 2008.
Understanding action beyond imitation: reversed compatibility effects of action
  observation in imitation and joint action.
\textit{Journal of Experimental Psychology: Human Perception and Performance}
  34:1493

\bibitem{poljac2009understanding}
Poljac E, van Schie HT, Bekkering H. 2009.
Understanding the flexibility of action--perception coupling.
\textit{Psychological Research PRPF} 73:578--586

\bibitem{galantucci2005experimental}
Galantucci B. 2005.
An experimental study of the emergence of human communication systems.
\textit{Cognitive science} 29:737--767

\bibitem{tomasello2005understanding}
Tomasello M, Carpenter M, Call J, Behne T, Moll H. 2005.
Understanding and sharing intentions: The origins of cultural cognition.
\textit{Behavioral and brain sciences} 28:675--691

\bibitem{wolpert1995internal}
Wolpert DM, Ghahramani Z, Jordan MI. 1995.
An internal model for sensorimotor integration.
\textit{Science} 269:1880

\bibitem{kilner07}
Kilner JM, Friston KJ, Frith CD. 2007.
Predictive coding: an account of the mirror neuron system.
\textit{Cognitive processing} 8:159--166

\bibitem{pecenka2011role}
Pecenka N, Keller PE. 2011.
The role of temporal prediction abilities in interpersonal sensorimotor
  synchronization.
\textit{Experimental Brain Research} 211:505--515

\bibitem{flanagan2003action}
Flanagan JR, Johansson RS. 2003.
Action plans used in action observation.
\textit{Nature} 424:769

\bibitem{koster2013theory}
Koster-Hale J, Saxe R. 2013.
Theory of mind: a neural prediction problem.
\textit{Neuron} 79:836--848

\bibitem{sartori2009does}
Sartori L, Becchio C, Bara BG, Castiello U. 2009.
Does the intention to communicate affect action kinematics?
\textit{Consciousness and cognition} 18:766--772

\bibitem{scorolli2014give}
Scorolli C, Miatton M, Wheaton LA, Borghi AM. 2014.
I give you a cup, i get a cup: a kinematic study on social intention.
\textit{Neuropsychologia} 57:196--204

\bibitem{candidi2015interactional}
Candidi M, Curioni A, Donnarumma F, Sacheli LM, Pezzulo G. 2015.
Interactional leader--follower sensorimotor communication strategies during
  repetitive joint actions.
\textit{Journal of The Royal Society Interface} 12:20150644

\bibitem{ho2015speaking}
Ho S, Foulsham T, Kingstone A. 2015.
Speaking and listening with the eyes: gaze signaling during dyadic
  interactions.
\textit{PloS one} 10:e0136905

\bibitem{frischen2007gaze}
Frischen A, Bayliss AP, Tipper SP. 2007.
Gaze cueing of attention: visual attention, social cognition, and individual
  differences.
\textit{Psychological bulletin} 133:694

\bibitem{huang2015using}
Huang CM, Andrist S, Saupp{\'e} A, Mutlu B. 2015.
Using gaze patterns to predict task intent in collaboration.
\textit{Frontiers in psychology} 6:1049

\bibitem{vesper2016role}
Vesper C, Schmitz L, Safra L, Sebanz N, Knoblich G. 2016.
The role of shared visual information for joint action coordination.
\textit{Cognition} 153:118--123

\bibitem{jack14}
Jack RE, Garrod OG, Schyns PG. 2014.
Dynamic facial expressions of emotion transmit an evolving hierarchy of signals
  over time.
\textit{Current biology} 24:187--192

\bibitem{hostetter2008visible}
Hostetter AB, Alibali MW. 2008.
Visible embodiment: Gestures as simulated action.
\textit{Psychonomic bulletin \& review} 15:495--514

\bibitem{goodrich2007human}
Goodrich MA, Schultz AC. 2007.
Human-robot interaction: a survey.
\textit{Foundations and trends in human-computer interaction} 1:203--275

\bibitem{kaspar14}
Kaspar K, K{\"o}nig S, Schwandt J, K{\"o}nig P. 2014.
The experience of new sensorimotor contingencies by sensory augmentation.
\textit{Consciousness and cognition} 28:47--63

\bibitem{MayeE13}
Maye A, Engel AK. 2013.
Extending sensorimotor contingency theory: prediction, planning, and action
  generation.
\textit{Adaptive Behaviour} 21:423--436

\bibitem{Saal10}
Saal HP, Ting J, Vijayakumar S. 2010.
Active sequential learning with tactile feedback. In \textit{Proceedings of the
  Thirteenth International Conference on Artificial Intelligence and Statistics
  (AISTATS-10)}, eds. YW~Teh, DM~Titterington, vol.~9

\bibitem{levine16}
Levine S, Finn C, Darrell T, Abbeel P. 2016.
End-to-end training of deep visuomotor policies.
\textit{Journal of Machine Learning Research} 17:1--40

\bibitem{ghadirzadeh2017deep}
Ghadirzadeh A, Maki A, Kragic D, Bj{\"o}rkman M. 2017.
Deep predictive policy training using reinforcement learning.
\textit{arXiv preprint arXiv:1703.00727}

\bibitem{ghadirzadeh16}
Ghadirzadeh A, B{\"u}tepage J, Kragic D, Bj{\"o}rkman M. 2016.
Self-learning and adaptation in a sensorimotor framework. In \textit{IEEE
  International Conference on Robotics and Automation, 2016. Proceedings.}

\bibitem{wahlstrom15}
Wahlstr{\"o}m N, Schoen TB, Deisenroth MP. 2015.
From pixels to torques: Policy learning with deep dynamical models.
\textit{arXiv preprint arXiv:1502.02251}

\bibitem{ghadirzadeh16b}
Ghadirzadeh A, B{\"u}tepage J, Maki A, Kragic D, Bj{\"o}rkman M. ????
A sensorimotor reinforcement learning framework for physical human-robot
  interaction. In \textit{Intelligent Robots and Systems (IROS), 2016 IEEE/RSJ
  International Conference on}. IEEE

\bibitem{ivaldi14}
Ivaldi S, Nguyen SM, Lyubova N, Droniou A, Padois V, et~al. 2014.
Object learning through active exploration.
\textit{IEEE Transactions on Autonomous Mental Development} 6:56--72

\bibitem{weinland11}
Weinland D, Ronfard R, Boyer E. 2011.
A survey of vision-based methods for action representation, segmentation and
  recognition.
\textit{Computer vision and image understanding} 115:224--241

\bibitem{nehaniv02}
Nehaniv CL, Dautenhahn K. 2002.
The correspondence problem. In \textit{Imitation in animals and artifacts}.
  41--61

\bibitem{mohammad13}
Mohammad Y, Nishida T. 2013.
Tackling the correspondence problem. In \textit{International Conference on
  Active Media Technology}. Springer

\bibitem{calinon07}
Calinon S, Billard A. 2007.
Incremental learning of gestures by imitation in a humanoid robot. In
  \textit{Proceedings of the ACM/IEEE international conference on Human-robot
  interaction}. ACM

\bibitem{shon05}
Shon AP, Grochow K, Rao RPN. 2005.
Robotic imitation from human motion capture using gaussian processes. In
  \textit{Humanoid Robots, 2005 5th IEEE-RAS}. IEEE

\bibitem{lei15}
Lei J, Song M, Li ZN, Chen C. 2015.
Whole-body humanoid robot imitation with pose similarity evaluation.
\textit{Signal Processing} 108:136--146

\bibitem{zuher12}
Zuher F, Romero R. 2012.
Recognition of human motions for imitation and control of a humanoid robot. In
  \textit{Robotics Symposium and Latin American Robotics Symposium (SBR-LARS),
  2012 Brazilian}. IEEE

\bibitem{schaal03}
Schaal S, Ijspeert A, Billard A. 2003.
Computational approaches to motor learning by imitation.
\textit{Philosophical Transactions of the Royal Society B: Biological Sciences}
  358:537--547

\bibitem{boucenna2014learning}
Boucenna S, Anzalone S, Tilmont E, Cohen D, Chetouani M. 2014.
Learning of social signatures through imitation game between a robot and a
  human partner.
\textit{IEEE Transactions on Autonomous Mental Development} 6:213--225

\bibitem{ijspeert2003learning}
Ijspeert AJ, Nakanishi J, Schaal S. 2003.
Learning attractor landscapes for learning motor primitives.
\textit{Advances in neural information processing systems} :1547--1554

\bibitem{amor2014interaction}
Amor HB, Neumann G, Kamthe S, Kroemer O, Peters J. 2014.
Interaction primitives for human-robot cooperation tasks. In \textit{Robotics
  and Automation (ICRA), 2014 IEEE International Conference on}. IEEE

\bibitem{gibson14}
Gibson JJ. 2014 (1. edition 1979.
The ecological approach to visual perception: classic edition.
Psychology Press

\bibitem{horton12}
Horton TE, Chakraborty A, Amant RS. 2012.
Affordances for robots: a brief survey.
\textit{Avant} 3:70--84

\bibitem{sun10}
Sun J, Moore JL, Bobick A, Rehg JM. 2010.
Learning visual object categories for robot affordance prediction.
\textit{The International Journal of Robotics Research} 29:174--197

\bibitem{chu16}
Chu V, Thomaz AL. 2016.
Learning and grounding haptic affordances using demonstration and human-guided
  exploration. In \textit{2016 11th ACM/IEEE International Conference on
  Human-Robot Interaction (HRI)}. IEEE

\bibitem{koppula16}
Koppula HS, Saxena A. 2016.
Anticipating human activities using object affordances for reactive robotic
  response.
\textit{IEEE transactions on pattern analysis and machine intelligence}
  38:14--29

\bibitem{kjellstrom11}
Kjellstr{\"o}m H, Romero J, Kragi{\'c} D. 2011.
Visual object-action recognition: Inferring object affordances from human
  demonstration.
\textit{Computer Vision and Image Understanding} 115:81--90

\bibitem{koppula13}
Koppula HS, Gupta R, Saxena A. 2013.
Learning human activities and object affordances from rgb-d videos.
\textit{The International Journal of Robotics Research} 32:951--970

\bibitem{lorenz11}
Lorenz T, M{\"o}rtl A, Vlaskamp B, Schub{\"o} A, Hirche S. 2011.
Synchronization in a goal-directed task: human movement coordination with each
  other and robotic partners. In \textit{RO-MAN, 2011 IEEE}. IEEE

\bibitem{mortl14}
M{\"o}rtl A, Lorenz T, Hirche S, Vasilaki E. 2014.
Rhythm patterns interaction-synchronization behavior for human-robot joint
  action.
\textit{PloS one} 9:e95195

\bibitem{ijspeert02}
Ijspeert AJ, Nakanishi J, Schaal S. 2002.
Learning rhythmic movements by demonstration using nonlinear oscillators. In
  \textit{Proceedings of the ieee/rsj int. conference on intelligent robots and
  systems (iros2002)}, no. BIOROB-CONF-2002-003

\bibitem{crick2006}
Crick C, Munz M, Scassellati B. 2006.
Synchronization in social tasks: Robotic drumming. In \textit{Robot and Human
  Interactive Communication, 2006. ROMAN 2006. The 15th IEEE International
  Symposium on}. IEEE

\bibitem{prepin2010agent}
Prepin K, Gaussier P. 2010.
How an agent can detect and use synchrony parameter of its own interaction with
  a human? In \textit{Development of Multimodal Interfaces: Active Listening
  and Synchrony}. Springer,  50--65

\bibitem{miyake2009interpersonal}
Miyake Y. 2009.
Interpersonal synchronization of body motion and the walk-mate walking support
  robot.
\textit{IEEE Transactions on Robotics} 25:638--644

\bibitem{broz12}
Broz F, Nehaniv CL, Kose-Bagci H, Dautenhahn K. 2012.
Interaction histories and short term memory: Enactive development of
  turn-taking behaviors in a childlike humanoid robot.
\textit{arXiv preprint arXiv:1202.5600}

\bibitem{leite13}
Leite I, Hajishirzi H, Andrist S, Lehman J. 2013.
Managing chaos: models of turn-taking in character-multichild interactions. In
  \textit{Proceedings of the 15th ACM on International conference on multimodal
  interaction}. ACM

\bibitem{calisgan12}
Calisgan E, Haddadi A, Van~der Loos HFM, Alcazar JA, Croft EA. 2012.
Identifying nonverbal cues for automated human-robot turn-taking. In
  \textit{RO-MAN, 2012 IEEE}. IEEE

\bibitem{hart14}
Hart JW, Gleeson B, Pan M, Moon A, MacLean K, Croft E. 2014.
Gesture, gaze, touch, and hesitation: Timing cues for collaborative work. In
  \textit{HRI Workshop on Timing in Human-Robot Interaction, Bielefeld,
  Germany}

\bibitem{skantze14}
Skantze G, Hjalmarsson A, Oertel C. 2014.
Turn-taking, feedback and joint attention in situated human--robot interaction.
\textit{Speech Communication} 65:50--66

\bibitem{okumura13}
Okumura Y, Kanakogi Y, Kanda T, Ishiguro H, Itakura S. 2013.
Infants understand the referential nature of human gaze but not robot gaze.
\textit{Journal of experimental child psychology} 116:86--95

\bibitem{admoni2011robot}
Admoni H, Bank C, Tan J, Toneva M, Scassellati B. 2011.
Robot gaze does not reflexively cue human attention. In \textit{CogSci}.
  Citeseer

\bibitem{moubayed2013}
MOUBAYED SA, Skantze G, Beskow J. 2013.
The furhat back-projected humanoid head--lip reading, gaze and multi-party
  interaction.
\textit{International Journal of Humanoid Robotics} 10:1350005

\bibitem{admoni2013you}
Admoni H, Hayes B, Feil-Seifer D, Ullman D, Scassellati B. 2013.
Are you looking at me?: perception of robot attention is mediated by gaze type
  and group size. In \textit{Proceedings of the 8th ACM/IEEE international
  conference on Human-robot interaction}. IEEE Press

\bibitem{doniec06}
Doniec MW, Sun G, Scassellati B. 2006.
Active learning of joint attention. In \textit{Humanoid Robots, 2006 6th
  IEEE-RAS}. IEEE

\bibitem{nagai06}
Nagai Y, Asada M, Hosoda K. 2006.
Learning for joint attention helped by functional development.
\textit{Advanced Robotics} 20:1165--1181

\bibitem{yucel13}
Yucel Z, Salah AA, Meri{\c{c}}li {\c{C}}, Meri{\c{c}}li T, Valenti R, Gevers T.
  2013.
Joint attention by gaze interpolation and saliency.
\textit{Cybernetics, IEEE Transactions on} 43:829--842

\bibitem{lawitzky2010}
Lawitzky M, M{\"o}rtl A, Hirche S. 2010.
Load sharing in human-robot cooperative manipulation. In \textit{19th
  International Symposium in Robot and Human Interactive Communication}. IEEE

\bibitem{evrard09}
Evrard P, Kheddar A. 2009.
Homotopy switching model for dyad haptic interaction in physical collaborative
  tasks. In \textit{EuroHaptics conference, 2009 and Symposium on Haptic
  Interfaces for Virtual Environment and Teleoperator Systems. World Haptics
  2009. Third Joint}. IEEE

\bibitem{kheddar11}
Kheddar A. 2011.
Human-robot haptic joint actions is an equal control-sharing approach possible?
  In \textit{Human System Interactions (HSI), 2011 4th International Conference
  on}. IEEE

\bibitem{mortl2012}
M{\"o}rtl A, Lawitzky M, Kucukyilmaz A, Sezgin M, Basdogan C, Hirche S. 2012.
The role of roles: Physical cooperation between humans and robots.
\textit{The International Journal of Robotics Research} 31:1656--1674

\bibitem{ikemoto2012physical}
Ikemoto S, Amor HB, Minato T, Jung B, Ishiguro H. 2012.
Physical human-robot interaction: Mutual learning and adaptation.
\textit{IEEE robotics \& automation magazine} 19:24--35

\bibitem{nikolaidis15}
Nikolaidis S, Ramakrishnan R, Gu K, Shah J. 2015.
Efficient model learning from joint-action demonstrations for human-robot
  collaborative tasks. In \textit{Proceedings of the Tenth Annual ACM/IEEE
  International Conference on Human-Robot Interaction}. ACM

\bibitem{nikolaidis2016formalizing}
Nikolaidis S, Kuznetsov A, Hsu D, Srinivasa S. 2016.
Formalizing human-robot mutual adaptation: A bounded memory model. In
  \textit{Human-Robot Interaction (HRI), 2016 11th ACM/IEEE International
  Conference on}. IEEE

\bibitem{thobbi11}
Thobbi A, Gu Y, Sheng W. 2011.
Using human motion estimation for human-robot cooperative manipulation. In
  \textit{Intelligent Robots and Systems (IROS), 2011 IEEE/RSJ International
  Conference on}. IEEE

\bibitem{kuderer12}
Kuderer M, Kretzschmar H, Sprunk C, Burgard W. 2012.
Feature-based prediction of trajectories for socially compliant navigation. In
  \textit{Proceedings of Robotics: Science and Systems}. Sydney, Australia

\bibitem{mainprice2013human}
Mainprice J, Berenson D. 2013.
Human-robot collaborative manipulation planning using early prediction of human
  motion. In \textit{2013 IEEE/RSJ International Conference on Intelligent
  Robots and Systems}. IEEE

\bibitem{micelli2011perception}
Micelli V, Strabala K, Srinivasa S. 2011.
Perception and control challenges for effective human-robot handoffs. In
  \textit{In Robotics: Science and systems workshop on rgb-d cameras}

\bibitem{prada2014implementation}
Prada M, Remazeilles A, Koene A, Endo S. 2014.
Implementation and experimental validation of dynamic movement primitives for
  object handover. In \textit{2014 IEEE/RSJ International Conference on
  Intelligent Robots and Systems}. IEEE

\bibitem{butepage2017deep}
B{\"u}tepage J, Black M, Kragic D, Kjellstr{\"o}m H. 2017.
Deep representation learning for human motion prediction and classification.
\textit{arXiv preprint arXiv:1702.07486}

\bibitem{butepage2017anticipating}
B{\"u}tepage J, Kjellstr{\"o}m H, Kragic D. 2017.
Anticipating many futures: Online human motion prediction and synthesis for
  human-robot collaboration.
\textit{arXiv preprint arXiv:1702.08212}

\bibitem{broz13}
Broz F, Nourbakhsh I, Simmons R. 2013.
Planning for human--robot interaction in socially situated tasks.
\textit{International Journal of Social Robotics} 5:193--214

\bibitem{wang13}
Wang Z, M{\"u}lling K, Deisenroth MP, Amor HB, Vogt D, et~al. 2013.
Probabilistic movement modeling for intention inference in human--robot
  interaction.
\textit{The International Journal of Robotics Research} 32:841--858

\bibitem{dragan2013legibility}
Dragan AD, Lee KC, Srinivasa SS. 2013.
Legibility and predictability of robot motion. In \textit{2013 8th ACM/IEEE
  International Conference on Human-Robot Interaction (HRI)}. IEEE

\bibitem{dragan2015effects}
Dragan AD, Bauman S, Forlizzi J, Srinivasa SS. 2015.
Effects of robot motion on human-robot collaboration. In \textit{Proceedings of
  the Tenth Annual ACM/IEEE International Conference on Human-Robot
  Interaction}. ACM

\bibitem{holladay2014legible}
Holladay RM, Dragan AD, Srinivasa SS. 2014.
Legible robot pointing. In \textit{The 23rd IEEE International Symposium on
  Robot and Human Interactive Communication}. IEEE

\bibitem{sciutti2014understanding}
Sciutti A, Patane L, Nori F, Sandini G. 2014.
Understanding object weight from human and humanoid lifting actions.
\textit{IEEE Transactions on Autonomous Mental Development} 6:80--92

\bibitem{cakmak2011using}
Cakmak M, Srinivasa SS, Lee MK, Kiesler S, Forlizzi J. 2011.
Using spatial and temporal contrast for fluent robot-human hand-overs. In
  \textit{Proceedings of the 6th international conference on Human-robot
  interaction}. ACM

\bibitem{kartoun2010human}
Kartoun U, Stern H, Edan Y. 2010.
A human-robot collaborative reinforcement learning algorithm.
\textit{Journal of Intelligent \& Robotic Systems} 60:217--239

\bibitem{medina2011experience}
Medina JR, Lawitzky M, M{\"o}rtl A, Lee D, Hirche S. 2011.
An experience-driven robotic assistant acquiring human knowledge to improve
  haptic cooperation. In \textit{Intelligent Robots and Systems (IROS), 2011
  IEEE/RSJ International Conference on}. IEEE

\bibitem{hayes2014discovering}
Hayes B, Scassellati B. 2014.
Discovering task constraints through observation and active learning. In
  \textit{Intelligent Robots and Systems (IROS 2014), 2014 IEEE/RSJ
  International Conference on}. IEEE

\bibitem{rozo2013learning}
Rozo L, Calinon S, Caldwell D, Jim{\'e}nez P, Torras C. 2013.
Learning collaborative impedance-based robot behaviors. In \textit{Proceedings
  of the Twenty-Seventh AAAI Conference on Artificial Intelligence}. AAAI Press

\bibitem{huang2016anticipatory}
Huang CM, Mutlu B. 2016.
Anticipatory robot control for efficient human-robot collaboration. In
  \textit{Human-Robot Interaction (HRI), 2016 11th ACM/IEEE International
  Conference on}. IEEE

\bibitem{huang2014learning}
Huang CM, Mutlu B. 2014.
Learning-based modeling of multimodal behaviors for humanlike robots. In
  \textit{Proceedings of the 2014 ACM/IEEE international conference on
  Human-robot interaction}. ACM

\bibitem{li2014human}
Li Y, Ge SS. 2014.
Human--robot collaboration based on motion intention estimation.
\textit{IEEE/ASME Transactions on Mechatronics} 19:1007--1014

\bibitem{whitney2016interpreting}
Whitney D, Eldon M, Oberlin J, Tellex S. 2016.
Interpreting multimodal referring expressions in real time. In \textit{Robotics
  and Automation (ICRA), 2016 IEEE International Conference on}. IEEE

\bibitem{ewerton2015learning}
Ewerton M, Neumann G, Lioutikov R, Amor HB, Peters J, Maeda G. 2015.
Learning multiple collaborative tasks with a mixture of interaction primitives.
  In \textit{Robotics and Automation (ICRA), 2015 IEEE International Conference
  on}. IEEE

\bibitem{sadigh2016information}
Sadigh D, Sastry SS, Seshia SA, Dragan A. 2016.
Information gathering actions over human internal state. In \textit{Intelligent
  Robots and Systems (IROS), 2016 IEEE/RSJ International Conference on}. IEEE

\bibitem{koppula2016anticipatory}
Koppula HS, Jain A, Saxena A. 2016.
Anticipatory planning for human-robot teams. In \textit{Experimental Robotics}.
  Springer

\bibitem{mainprice2010planning}
Mainprice J, Sisbot EA, Sim{\'e}on T, Alami R. 2010.
Planning safe and legible hand-over motions for human-robot interaction. In
  \textit{IARP workshop on technical challenges for dependable robots in human
  environments}, vol.~2

\bibitem{mainprice2012sharing}
Mainprice J, Gharbi M, Sim{\'e}on T, Alami R. 2012.
Sharing effort in planning human-robot handover tasks. In \textit{RO-MAN, 2012
  IEEE}. IEEE

\bibitem{nikolaidis2013human}
Nikolaidis S, Shah J. 2013.
Human-robot cross-training: computational formulation, modeling and evaluation
  of a human team training strategy. In \textit{Proceedings of the 8th ACM/IEEE
  international conference on Human-robot interaction}. IEEE Press

\bibitem{mutlu2013coordination}
Mutlu B, Terrell A, Huang CM. 2013.
Coordination mechanisms in human-robot collaboration. In \textit{Proceedings of
  the Workshop on Collaborative Manipulation, 8th ACM/IEEE International
  Conference on Human-Robot Interaction}. Citeseer

\bibitem{young2011evaluating}
Young JE, Sung J, Voida A, Sharlin E, Igarashi T, et~al. 2011.
Evaluating human-robot interaction.
\textit{International Journal of Social Robotics} 3:53--67

\bibitem{giuliani2010design}
Giuliani M, Lenz C, M{\"u}ller T, Rickert M, Knoll A. 2010.
Design principles for safety in human-robot interaction.
\textit{International Journal of Social Robotics} 2:253--274

\bibitem{hoffman2007effects}
Hoffman G, Breazeal C. 2007.
Effects of anticipatory action on human-robot teamwork efficiency, fluency, and
  perception of team. In \textit{Proceedings of the ACM/IEEE international
  conference on Human-robot interaction}. ACM

\bibitem{leite2015comparing}
Leite I, McCoy M, Ullman D, Salomons N, Scassellati B. 2015.
Comparing models of disengagement in individual and group interactions. In
  \textit{Proceedings of the Tenth Annual ACM/IEEE International Conference on
  Human-Robot Interaction}. ACM

\bibitem{rolf2009attention}
Rolf M, Hanheide M, Rohlfing KJ. 2009.
Attention via synchrony: Making use of multimodal cues in social learning.
\textit{IEEE Transactions on Autonomous Mental Development} 1:55--67

\bibitem{meisner2009shadowplay}
Meisner EM, {\`A}banovic S, Isler V, Caporeal LR, Trinkle J. 2009.
Shadowplay: a generative model for nonverbal human-robot interaction. In
  \textit{Proceedings of the 4th ACM/IEEE international conference on Human
  robot interaction}. ACM

\bibitem{sandoval2016reciprocity}
Sandoval EB, Brandstetter J, Obaid M, Bartneck C. 2016.
Reciprocity in human-robot interaction: a quantitative approach through the
  prisoner’s dilemma and the ultimatum game.
\textit{International Journal of Social Robotics} 8:303--317

\bibitem{harriott2015measuring}
Harriott CE, Zhang T, Buford GL, Adams JA. 2015.
Measuring human workload in a collaborative human-robot team.
\textit{Journal of Human-Robot Interaction} 4:61--96

\bibitem{bartneck2009measurement}
Bartneck C, Kuli{\'c} D, Croft E, Zoghbi S. 2009.
Measurement instruments for the anthropomorphism, animacy, likeability,
  perceived intelligence, and perceived safety of robots.
\textit{International journal of social robotics} 1:71--81

\bibitem{warneken2012collaborative}
Warneken F, Gr{\"a}fenhain M, Tomasello M. 2012.
Collaborative partner or social tool? new evidence for young children’s
  understanding of joint intentions in collaborative activities.
\textit{Developmental science} 15:54--61

\end{thebibliography}

\end{document}